\documentclass[journal]{IEEEtran}

\IEEEoverridecommandlockouts
\usepackage{graphicx}
\def\BibTeX{{\rm B\kern-.05em{\sc i\kern-.025em b}\kern-.08em
    T\kern-.1667em\lower.7ex\hbox{E}\kern-.125emX}}
\usepackage[font=small,labelfont=bf,tableposition=top]{caption}

\usepackage{blindtext}
\usepackage{caption}

\let\oldtwocolumn\twocolumn
\renewcommand\twocolumn[1][]{%
    \oldtwocolumn[{#1}{
    \vskip-5ex
        \centering
        \includegraphics[width=0.99\textwidth]{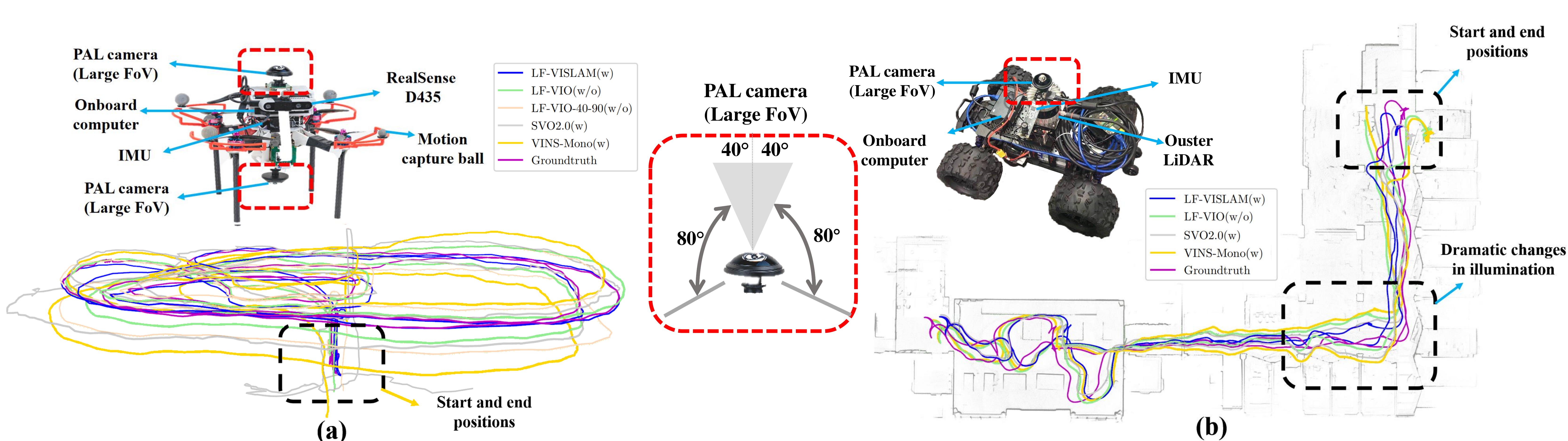}
        \captionof{figure} {Our proposed LF-VISLAM is designed for large-FoV cameras, which is suitable for mobile agents such as aerial- and ground vehicles. \textbf{(a)} On the top is our flight experiment platform with two Panoramic Annular Lens (PAL) cameras, a RealSense D435 sensor, a flight control system with an IMU sensor, and an onboard computer. {On the bottom are trajectory results of different SLAM systems on our PALVIO dataset sequence ID01}. \textbf{(b)} On the top is our car experiment platform with a PAL camera, an IMU sensor, an Ouster LiDAR, and an onboard computer. {On the bottom are long trajectory results of different SLAM systems on our PALVIO dataset sequence IDL01.} Our method LF-VISLAM not only provides better localization accuracy but also has wide applicability in various scenarios for mobile agents.}
        \label{fig:head_image}
    }]
}

\usepackage{ifpdf}
\usepackage{cite}
\usepackage{amsmath}
\usepackage{array}
\usepackage{fixltx2e}
\usepackage{dblfloatfix}
\usepackage{url}

\usepackage{float}
\usepackage{booktabs}
\usepackage{multirow}
\usepackage{cite}
\usepackage{amsmath}
\usepackage{bbding} 
\usepackage[colorlinks,linkcolor=red,anchorcolor=blue,urlcolor=magenta,citecolor=blue]{hyperref} 
\usepackage[ruled]{algorithm2e}
\usepackage{graphicx}
\usepackage{xspace}
\usepackage{amsfonts}

\usepackage{xcolor}
\usepackage{pifont}  
\usepackage{fancyref}
\usepackage{booktabs}
\usepackage{threeparttable}  

\usepackage{subfigure}

\graphicspath{{imgs}{old_imgs/}{old_imgs_1/}}

\newcommand{\red}[1]{{\color{red}{#1}}}

\newcommand{\blue}[1]{{\color{blue}{#1}}}
\newcommand{\orange}[1]{\textcolor[RGB]{255,192,0}{#1}}

\newcommand{\green}[1]{\textcolor[RGB]{0,176,80}{#1}}

\definecolor{rblue}{rgb}{0,0.5,1}

\begin{document}
\title{LF-VISLAM: A SLAM Framework for Large Field-of-View Cameras with Negative\\Imaging Plane on Mobile Agents}

\author{Ze Wang$^{\dag,1,5}$, Kailun Yang$^{\dag,2,3}$, Hao Shi$^{1}$, Peng Li$^{1}$, Fei Gao$^{4,5}$, Jian Bai$^{1}$, and Kaiwei Wang$^{1}$%
\thanks{This was supported in part by the National Natural Science Foundation of China (Grant No. 12174341), in part by the National Key R\&D Program of China (Grant No. 2022YFF0705500), in part by Hangzhou SurImage Technology Co. Ltd., and in part by Hangzhou HuanJun Technology Co. Ltd.
{\textit{(Corresponding authors: Kaiwei Wang and Kailun Yang.)}}
}%
\thanks{${\dag}$denotes equal contribution.}
\thanks{$^{1}$State Key Laboratory of Extreme Photonics and Instrumentation, Zhejiang University, China}
\thanks{$^{2}$School of Robotics, Hunan University, China}
\thanks{$^{3}$National Engineering Research Center of Robot Visual Perception and Control Technology, Hunan University, China}
\thanks{$^{4}$State Key Laboratory of Industrial Control Technology, Zhejiang University, China}
\thanks{$^{5}$Huzhou Institute of Zhejiang University, Zhejiang University, China.}
\thanks{Email: \{wangze0527, haoshi, peng\_li, fgaoaa, bai, wangkaiwei\}@zju.edu.cn, kailun.yang@hnu.edu.cn.}
}

\markboth{IEEE Transactions on Automation Science and Engineering, October 2023}%
{Wang \MakeLowercase{\textit{et al.}}: LF-VISLAM}

\maketitle

\begin{abstract}
Simultaneous Localization And Mapping (SLAM) has become a crucial aspect in the fields of autonomous driving and robotics. One crucial component of visual SLAM is the Field-of-View (FoV) of the camera, as a larger FoV allows for a wider range of surrounding elements and features to be perceived. {However, when the FoV of the camera reaches the negative half-plane, traditional methods for representing image feature points using $\begin{bmatrix}u,v,1\end{bmatrix}^T$ become ineffective.} While the panoramic FoV is advantageous for loop closure, its benefits are not easily realized under large-attitude-angle differences where loop-closure frames cannot be easily matched by existing methods. As loop closure on wide-FoV panoramic data further comes with a large number of outliers, traditional outlier rejection methods are not directly applicable. {To address these issues, we propose LF-VISLAM, a \textbf{V}isual \textbf{I}nertial \textbf{SLAM} framework for cameras with extremely \textbf{L}arge \textbf{F}oV with loop closure. A three-dimensional vector with unit length is introduced to effectively represent feature points even on the negative half-plane.} The attitude information of the SLAM system is leveraged to guide the feature point detection of the loop closure. Additionally, a new outlier rejection method based on the unit length representation is integrated into the loop closure module. {We collect the PALVIO dataset using a \textbf{P}anoramic \textbf{A}nnular \textbf{L}ens (PAL) system with an entire FoV of $360^\circ{\times}(40^\circ{\sim}120^\circ)$ and an Inertial Measurement Unit (IMU) for \textbf{V}isual \textbf{I}nertial \textbf{O}dometry (VIO) to address the lack of panoramic SLAM datasets.} Experiments on the established PALVIO and public datasets show that the proposed LF-VISLAM outperforms state-of-the-art SLAM methods. Our code will be open-sourced at \url{https://github.com/flysoaryun/LF-VISLAM}.

\end{abstract}

\par
{\fontsize{10}{12}\selectfont 
\textbf{\emph{Note to Practitioners}— Motivated by the challenges of handling large-FoV cameras in SLAM applications, {this paper proposes LF-VISLAM, a novel SLAM framework that uses a large-FoV camera and IMU sensors.
Our framework is equipped with a loop closure thread that can use attitude information to eliminate accumulated errors. We have made algorithmic adjustments and optimizations to the negative half-plane features to better adapt to cameras with large FoV.
Experimental evaluations demonstrate that LF-VISLAM significantly outperforms traditional SLAM methods.} Additionally, the code will be open-sourced, providing easy access for research and implementation. Overall, LF-VISLAM is a promising solution to improve the performance of SLAM in challenging environments of autonomous driving and robotics.}


\IEEEpeerreviewmaketitle

\section{Introduction}
\IEEEPARstart
{T}HE advancement of robotics and autonomous vehicles has led to a growing application of SLAM in mobile navigation systems~\cite{chen2019palvo,hu2019indoor,seok2019rovo,seok2020rovins,sun2020plane,liang2019salientdso,yang2016monocular,li2021bridging}.
The use of cameras with a large Field-of-View (FoV) has also become increasingly common in these systems, \textit{e.g.}, aerial- and ground vehicles in Fig.~\ref{fig:head_image}, as they allow for sensing a wider range of elements and features in the surrounding environment, which can be beneficial for higher-level vision perception and odometry tasks~\cite{qian2022survey,yang2020dspass,lin2018pvo,gao2021lovins,chen2021semantic,jaus2021panoramic,shi2022panoflow}.
Some modern panoramic cameras~\cite{sun2019multimodal,yang2019pass,shan2021lvi} even have a negative plane field, which enables ultra-wide understanding of the surrounding environment, where imaging points can appear on the negative plane ($z\textless 0$, see Fig.~\ref{fig:negative_plane}).

\begin{figure}
	\centering
	\includegraphics[width=1.0\linewidth]{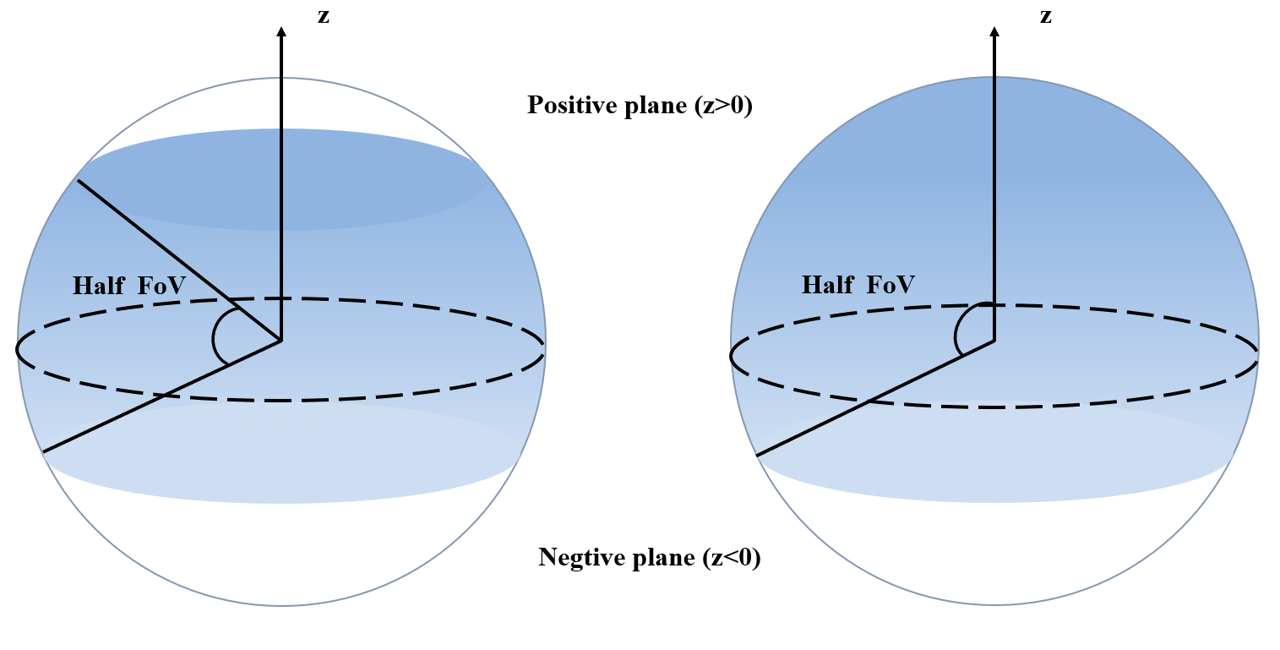}
	\caption{The left subfigure shows the half FoV of a panoramic annular camera or a catadioptric camera, and the right one shows the half FoV of a fisheye camera. The upper half above the dotted circle is the positive plane, while the bottom half is the negative plane.}
	\label{fig:negative_plane}
\end{figure}

Currently, there are various SLAM frameworks~\cite{forster2014svo,qin2018vins,campos2021orb} that can accommodate different types of camera models, such as pinhole, fisheye, and the omnidirectional camera model proposed by Scaramuzza~\textit{et al.}~\cite{scaramuzza2006toolbox}.
{Taking VINS-Mono~\cite{qin2018vins}, a SLAM system based on optimization, as an example, in both front-end and back-end of the visual processing, the entire framework uses $\begin{bmatrix}u,v,1\end{bmatrix}^T$ to represent the location of image feature points in the homogeneous coordinates.
In this context, the variables $u$ and $v$ represent the pixel coordinates of a point in the image plane.
The value $1$ serves as a scale factor.
The system itself does not take into account the existence of a complex plane and a camera with a large FoV.}
{Most existing systems~\cite{forster2014svo,qin2018vins,campos2021orb} follow the same strategy by directly discarding points within the negative half-plane.}
This poses a significant issue as this negative semi-planar region is present in various large-FoV systems such as fisheye, panoramic annular, and catadioptric cameras, and discarding these features results in the loss of important information. Near $180^\circ$, the values of $u$ and $v$ increase rapidly, which negatively affects algorithms such as PnP and epipolar constraints used to solve the rotation matrix and translation vector, resulting in reduced tracking accuracy and even failures.

Another challenge lies in the loop closure detection algorithm for panoramic frames.
{Loop closure refers to the detection and correction of errors in the estimated trajectory when the robot revisits a previously visited location. It is an important aspect of SLAM algorithms to improve accuracy and consistency.} {Existing panoramic Visual Inertial Odometry (VIO)~\cite{seok2020rovins} or Visual Odometry (VO)~\cite{chen2019palvo,seok2019rovo,lin2018pvo,wang2018cubemapslam,wang2022pal_slam,huang2022360vo} frameworks do not incorporate a loop closure module.}
While the panoramic FoV can be beneficial for loop closure, its advantages are limited under significant attitude differences, where loop-closure frames are difficult to match using existing methods~\cite{qin2018vins,forster2014svo}. Yet, the full potential of the ultra-wide FoV cannot be realized without a proper loop closure design.
{Further taking VINS-Mono~\cite{qin2018vins} for example, when the difference between the yaw angle and the tilt angle of two frames is very large, even if it is the same position, the loop-closure frames still cannot be recognized.}

To address these issues, we propose \emph{LF-VISLAM}, a SLAM framework for cameras with large FoV. 
Instead of using traditional representations, we propose to utilize a feature point vector with unit length to represent the features even on the negative plane (Sec.~\ref{sec:Initialization}).
We then introduce a sliding-window-based tightly-coupled monocular odometry for state estimation under the unit vector representation (Sec.~\ref{sec:Odometry}).
An improved loop closure detection method is seamlessly integrated into the LF-VISLAM framework to effectively incorporate large-FoV attitude guidance and further boost the success rate of descriptor matching (Sec.~\ref{sec:loop}).
{The outliers are finally identified using our proposed Efficient Perspective-n-Point (EPnP) RANdom SAmple Consensus (RANSAC) method to take into account the negative imaging plane case when checking the inlier unit vector feature points (Sec.~\ref{sec:epnp_ransac_unit_vector_feature_points})}.

Specifically, during initialization, epipolar geometry is used to initialize the system when there is sufficient parallax between two frames.
Additionally, feature extraction is performed on the raw panoramic image, bypassing the need for complex panorama unfolding and cropping into multiple pinhole sub-images, making our method fast and suitable for real-time mobile robotic agents.
The essential matrix~\cite{hartley1995investigation} is then decomposed into a rotation matrix and translation vector, and the correct values are selected.
Triangulation and EPnP alternation methods~\cite{lepetit2009epnp} are used to initialize the depth of feature points and all poses within the sliding window, and a tightly-coupled optimization method is applied to re-solve all the rotation and translation matrices in the sliding window.
Further, using the sliding window method to maintain the state and map information of the robot greatly shortens the time for SLAM to solve the optimization.

After the vision initialization, the IMU data and image data are aligned to recover the scale information.
In our approach, the optimization problem is solved by taking into account the visual re-projection error, the IMU pre-integration error, and the marginalization error. 
Given that the SLAM system can obtain attitude information during the loop closure process, we put forward to leverage this information to enhance the efficiency and accuracy of the loop closure, which allows for robust matching of frames, even when there is a large difference in tilt- and yaw angle, leading to an improvement in the accuracy of the loop closure, as shown in Fig.~\ref{fig:introduction}.
The Yaw angle refers to the angle between the projection of the body coordinate system's x-axis onto the x-O-y plane of the world coordinate system and the world coordinate system's x-axis. The z-axis of the world coordinate system is opposite to the direction of gravity. The Tilt angle represents the angle between the body coordinate system's z-axis and the world coordinate system's z-axis.

\begin{figure}[!t]

\begin{minipage}[!t]{1\linewidth}
  \centering
  \includegraphics[width=8cm]{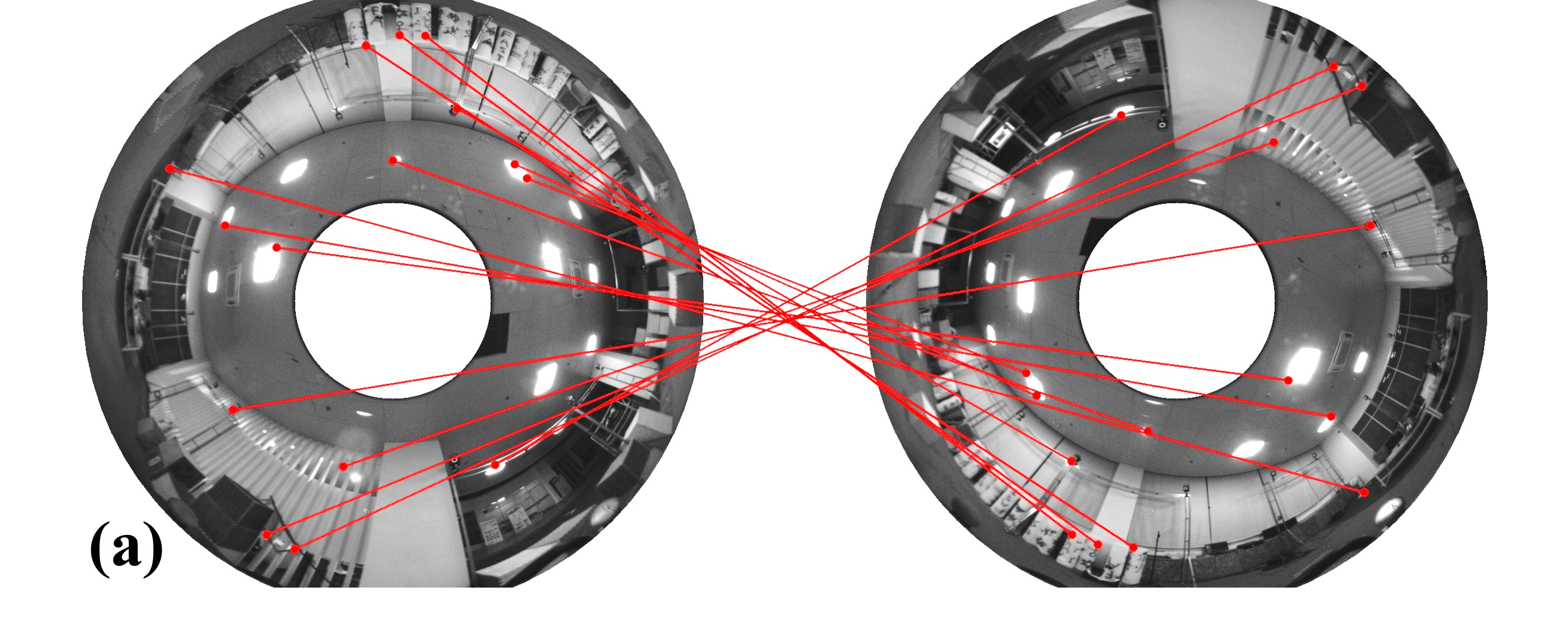}\\
\centering
 \end{minipage}%
 
\begin{minipage}[t]{1\linewidth}
  \centering
  \includegraphics[width=8cm]{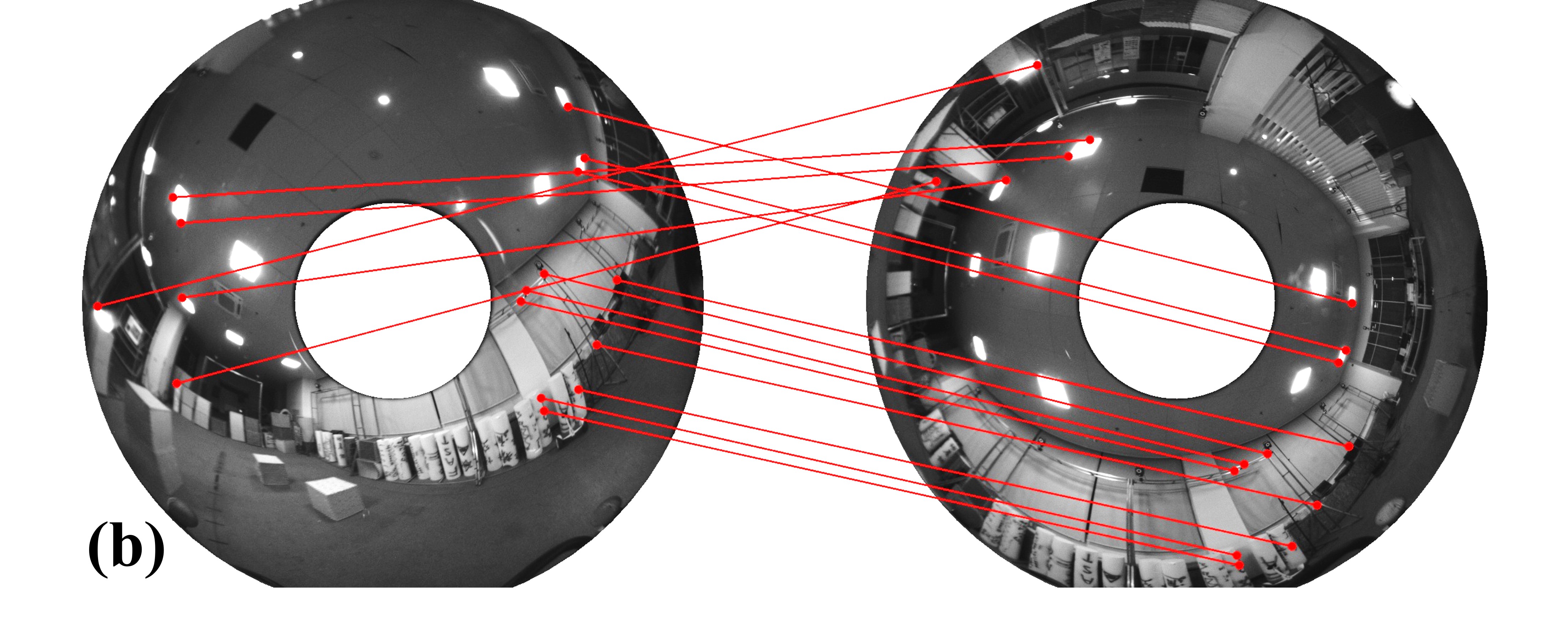}\\
\centering
\end{minipage}%
  \caption{Our loop images with large attitude differences (for illustration purposes, the illumination and contrast of the panoramas are improved) in scenarios \textbf{(a)} large yaw angle difference and \textbf{(b)} large tilt angle difference.\label{fig:introduction}
  }
\end{figure}

{Although there are many open-source datasets~\cite{geiger2013vision, burri2016euroc,schubert2018tum}, there is a lack of large-FoV camera-based databases for visual odometry.}
We introduce a new panoramic indoor visual SLAM dataset, termed as the \emph{PALVIO} dataset, to overcome the limited availability of panoramic visual odometry datasets with ground-truth location and pose, which is collected using a Panoramic Annular Lens (PAL) system with a full FoV of $360^\circ{\times}(40^\circ{\sim}120^\circ)$, an IMU sensor, and a motion capture device (refer to Fig.~\ref{fig:head_image}~\textbf{(a)}).
We conduct extensive experiments to evaluate our proposed LF-VISLAM framework on both the established PALVIO benchmark and a public fisheye camera dataset~\cite{shan2021lvi} with a FoV of $360^\circ{\times}(0^\circ{\sim}93.5^\circ)$.
Our investigations on different FoVs demonstrate the importance of the negative-plane information for an SLAM system, and the proposed method outperforms state-of-the-art VIO and SLAM frameworks. Moreover, we show that our method is beneficial when integrated with a LiDAR-visual-inertial odometry~\cite{shan2021lvi}.

In summary, we deliver the following contributions:
\begin{itemize}
    \item We propose LF-VISLAM, a SLAM framework for cameras with a large FoV, which includes a new method for robust initialization and a method to assist the descriptor extraction with VIO attitude information. In LF-VISLAM, a novel RANSAC method is designed for outlier rejection during pose transformation estimation.
    \item We create and release the PALVIO dataset, the first dataset for evaluating vehicles using cameras with a large field-of-view and IMU sensors. The dataset includes ground-truth location and pose obtained via a motion capture device.
    \item We experimentally validate and compare LF-VISLAM with conventional methods using both ground and aerial vehicles. LF-VISLAM outperforms state-of-the-art VIO and SLAM methods on multiple wide-FoV datasets and experiment sequences with a negative plane.
\end{itemize}

{This article extends our conference work~\cite{wang2022lfvio} with the following content:} 
\begin{itemize}
    \item {Extension of large-FoV visual-inertial-odometry: We present a complete large-FoV SLAM system that integrates loop closure blocks, enhancing overall performance by incorporating loop closure information.}
    \item {Innovative loop closure design: Our work introduces a novel approach to improve descriptor extraction in loop closure detection. We leverage VIO attitude information and propose a RANSAC-based method to reject outlier points, enabling accurate pose transformation estimation between loop-closure frames and keyframes.}
    \item  {Extended experimental validation: We evaluate our system on long-trajectory sequences, demonstrating its suitability for mobile navigation agents like aerial and ground vehicles. This validation showcases the practical applicability of our system in real-world scenarios.}
\end{itemize}

\section{Related Work}
In this section, a brief review of representative works is presented on small-FoV SLAM and panoramic Simultaneous Localization And Mapping (SLAM) frameworks.

\subsection{Narrow-FoV SLAM}
SLAM is the process of estimating the 3D pose (\textit{i.e.}, local position and orientation) and velocity relative to a local starting position, using various input signals like visual- and IMU data. 
{Qin~\textit{et al.}~\cite{qin2018vins} proposed VINS, a robust and versatile monocular vision inertial state estimator.}
Recently, the Semi-direct Visual-inertial Odometry (SVO2.0)~\cite{forster2014svo} has been introduced for monocular and multi-camera systems.
{Campos~\textit{et al.}~\cite{campos2021orb} proposed ORB-SLAM3, an accurate open-source library for vision, visual-inertial, and multi-map SLAM.}
LiDAR-visual-inertial sensor fusion frameworks have also been developed in recent years such as LVI-SAM~\cite{shan2021lvi} and R3LIVE~\cite{lin2021r3live}.
In addition to feature-point-based methods, there are also direct methods for VO and VIO which are more sensitive to light than feature-point-based methods.
Engel~\textit{et al.}~\cite{engel2017dso} proposed Direct Sparse Odometry (DSO) using a fully direct probabilistic model and Wang~\textit{et al.}~\cite{wang2017stereo_dso} combined it with inertial systems and stereo cameras.
{The above algorithms only consider cameras with a narrow FoV. For cameras with a large FoV, only the positive half-plane information can be used to solve the SLAM problem.}

Overall, SLAM frameworks have been broadly investigated in recent years, and various methods have been proposed to improve their accuracy, robustness, and real-time performance~\cite{yin2022dynam}.
Yet, it is worth noting that a large part of these methods are designed for monocular cameras, while others are designed for multi-camera systems, stereo cameras, or LiDAR-camera fusion systems~\cite{chou2021efficient}.
Differing from these existing methods, we propose LF-VISLAM, a universal SLAM system designed for large-FoV cameras like panoramic annular and fisheye cameras.
{In particular, LF-VISLAM is a complete SLAM system, which takes the ultra-wide viewing angle of these omnidirectional sensors into consideration and makes use of the feature points appearing on the negative imaging plane with an attitude-guided loop closure design.}

\begin{figure*}[!t]
	\centering
	\includegraphics[width=1.0\linewidth]{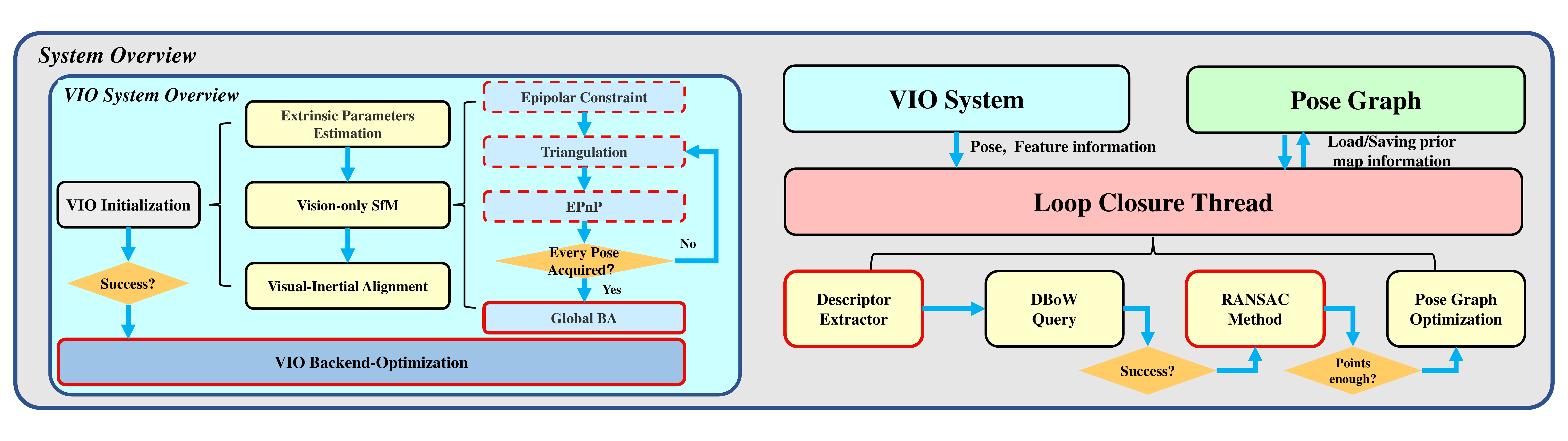}
	\caption{{Overview of our SLAM system with all blocks online. The left part shows the VIO system and the right part shows the whole system flow. Among them, the red solid line box represents our proposed method, and the red dotted line box is the modified part according to the application. The VIO system will pass the solved pose information (attitude and position) and feature point information (identity (ID), position in the image, and 3D position) to the loop closure thread.}}
	\label{fig:flow}
\end{figure*}

\subsection{Panoramic SLAM}
{
Panoramic cameras can help SLAM systems deliver better robustness in environments with weak textures due to their wider sensing range and the captured richer features~\cite{gao2022review,yang2021capturing,jaus2021panoramic}.
Therefore, there are many visual odometry, visual-inertial odometry, and SLAM frameworks based on panoramic cameras.
}
Sumikura~\textit{et al.}~\cite{sumikura2019openvslam} proposed OpenVSLAM, which is a VO framework supporting panoramic cameras.
Wang~\textit{et al.}~\cite{wang2018cubemapslam} put forward CubemapSLAM, a piecewise-pinhole monocular fisheye-image-based visual SLAM system.
Chen~\textit{et al.}~\cite{chen2019palvo,chen2021pa_slam} tackled visual odometry with a panoramic annular camera.
More recently, Wang~\textit{et al.}~\cite{wang2022pal_slam} proposed PAL-SLAM, which works with panoramic annular cameras.
Huang~\textit{et al.}~\cite{huang2022360vo} proposed 360VO, a direct visual odometry algorithm.

Ahmadi~\textit{et al.}~\cite{ahmadi2023hdpv} presented HDPV-SLAM, a novel visual SLAM method that uses a panoramic camera and a tilted multi-beam LiDAR scanner to generate accurate and metrically-scaled vehicle trajectories.
Yuwen~\textit{et al.}~\cite{yuwen2022gaze} proposed a method for actively improving the positioning accuracy of visual SLAM by controlling the gaze of a positioning camera mounted on an autonomous guided vehicle using a panoramic cost map.
Liu~\textit{et al.}~\cite{liu2022_360st_mapping} presented a novel online topological mapping method, 360ST-Mapping, which leverages omnidirectional vision and semantic information to incrementally extract and improve the representation of the environment.
Karpyshev~\textit{et al.}~\cite{karpyshev2022mucaslam} introduced the MuCaSLAM for improving the computational efficiency and robustness of visual SLAM algorithms on mobile robots using a neural network to predict whether the current image contains enough ORB features that can be matched with subsequent frames.

{
However, the above and many current visual odometry and visual-inertial odometry frameworks do not have modules for loop closure~\cite{won2020omnislam,matsuki2018omnidirectional_dso,ramezani2018pose_estimation_omnidirectional}, whereas some panoramic SLAM frameworks~\cite{wang2018cubemapslam,2021Omni} convert panoramic images into multiple pinhole images, which can be processed in parallel via a multi-thread program, but this paradigm sacrifices the $360^\circ$ consistency of the panoramic images.
}
In this work, we propose a large-FoV-oriented SLAM framework with loop closure directly performed on panoramas to fully materialize the benefits of $360^\circ$ visual content.

\section{LF-VISLAM: Proposed Framework}\label{sec:Method}

In this section, we explain in detail our proposed LF-VISLAM framework for large-FoV cameras with a negative plane, as shown in Fig.~\ref{fig:flow}. The LF-VISLAM framework mainly comprises two independent threads: the Visual-Inertial-Odometry (VIO) system and the loop closure. 
The VIO system mainly consists of two parts: the initialization described in Sec.~\ref{sec:Initialization} and the backend optimization detailed in Sec.~\ref{sec:Odometry}.
The VIO system provides pose and position information, as well as 3D coordinates of feature points, to the loop closure method presented in Sec.~\ref{sec:loop}.
The components denoted via red solid line boxes in the VIO system bring the negative half-plane information into the optimization.
The red dot line in the VIO system denotes the adjustments of traditional algorithms to introduce negative half-plane information.
The loop closure consists of feature descriptor extraction, DBoW query, RANSAC method, and pose graph optimization as described in Sec.~\ref{sec:epnp_ransac_unit_vector_feature_points}. We now elaborate on each component in the proposed LF-VISLAM. {The red solid line box in loop closure improves loop matching performance and improves robustness through outlier removal.}

\subsection{Initialization}
\label{sec:Initialization}

Our optimization approach to the SLAM problem leverages proper initial values, which are crucial for visual-inertial systems due to their non-convexity. 

The camera model serves as a bridge from pixel coordinates to the physical world. Therefore, the choice of the camera model is important.
As an example, pinhole cameras are naturally unable to characterize the negative half-plane of the camera due to the limitations of their models.
The panoramic optical system~\cite{chen2019palvo} utilizes refraction- or reflection-based optical imaging principles, giving it a large field of view.
Therefore, a more appropriate camera model such as the model introduced by Scaramuzza~\textit{et al.}~\cite{scaramuzza2006toolbox} is required for large field-of-view cameras, especially for those which can observe the camera's rear field-of-view.
For an imaging system with a negative half-plane field-of-view, to utilize its full image information, one needs to describe the pixel coordinate system by projecting it onto a spherical model, where the internal reference required for the projection can be obtained via checkerboard calibration~\cite{scaramuzza2006toolbox}.
Each pixel point $\mathbf{u}_i\in\mathbb{R}^2$ in the pixel coordinate system thus corresponds to a vector pointing from the center of the sphere to the surface of the unit sphere $\mathbf{w}_i\in\mathbb{S}^2$.
The point in unit sphere $\mathbf{w}_i$ can be obtained via Eq.~(\ref{eq:point_unit_1})(\ref{eq:point_unit_2}):
\begin{equation}
\label{eq:point_unit_1}
\mathbf{w}=\frac{\mathbf{v}}{||\mathbf{v}||_2},
\end{equation}
\begin{equation}
\label{eq:point_unit_2}
\mathbf{v}=\begin{bmatrix} x_i\\y_i\\a_0+a_1\rho+a_2\rho^2+...a_n\rho^ N\end{bmatrix}, \\
\rho=\sqrt{x_i^2+y_i^2},    
\end{equation}
where the coefficients $a_i\left(i=1,2...N\right)$ are the polynomial parameters and $x_i,y_i$ denotes the pixel point position. The polynomial coefficients are obtained through a calibration process using a tool such as the OmniCalib calibration toolbox~\cite{scaramuzza2006toolbox} specifically designed for panoramic cameras.

\begin{figure}
	\centering
	\includegraphics[width=1.0\linewidth]{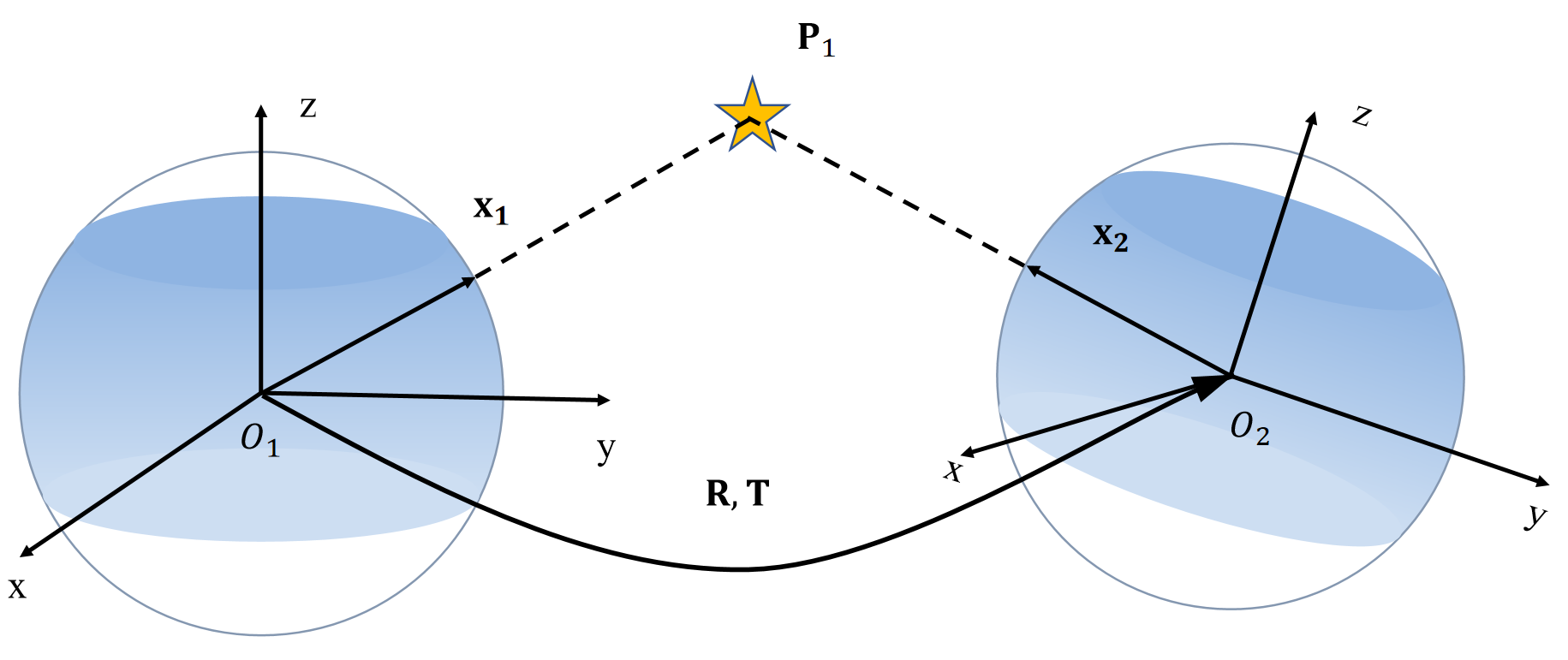}
	\caption{The epipolar constraint. The vectors of two corresponding features and their rotation and translation are on the same plane.}
	\label{fig:epipolar}
\end{figure}

We propose to characterize the feature point vectors $\begin{bmatrix}\alpha,\beta,\gamma\end{bmatrix}^T$ using vectors with $\alpha^2+\beta^2+\gamma^2=1$ constraints, and such vectors can represent points in the negative half-plane of the camera coordinate system.
Other camera models such as MEI~\cite{mei2007single} and Kannala-Brandt~\cite{kannala2006generic} can also be obtained as a mapping relationship from pixel points to unit sphere vectors.
In our experiments on the PALVIO dataset and long trajectories, the model from Scaramuzza~\textit{et al.}~\cite{scaramuzza2006toolbox} is chosen for efficiency and distortion considerations.
Algorithms must also be adjusted to work with cameras that have a negative plane. In our framework, we leverage Shi-Tomasi corners~\cite{shi1994good} and the Lucas-Kanade method~\cite{lucas1981iterative} to extract and track these corners. The polar geometry RANSAC method is used to eliminate the outlier points, for which the constraint is formulated as shown in Eq.~(\ref{eq:1})(\ref{eq:unit_vectors}):
\begin{equation}
\label{eq:1}
{
 {\mathbf{x}_2^T\left( \left[\mathbf{t}^{c_j}_{c_i}\right]^\wedge \mathbf{R}^{c_j}_{c_i} \right)\mathbf{x}_1=0}
},
\end{equation}
\begin{equation}
\label{eq:unit_vectors}
\mathbf{x}_1=\begin{bmatrix} \alpha_1 \\\beta_1 \\\gamma_1\end{bmatrix},\mathbf{x}_2=\begin{bmatrix} \alpha_2 \\\beta_2 \\\gamma_2\end{bmatrix},
\end{equation}
where $\mathbf{x}_1$ and $\mathbf{x}_2$ are the unit vectors corresponding to the same space point $\mathbf{P}_1$ in Fig.~\ref{fig:epipolar}, {$O_1$ is the center of the camera corresponding to the $i$-th frame, $O_2$ is the center of the camera corresponding to the $j$-th frame,} and the notation $[\mathbf{t}^{c_j}_{c_i}]^{\wedge}$ denotes the skew-symmetric cross product matrix of $\mathbf{t}^{c_j}_{c_i}\in{\mathbb{R}^3}$.

\noindent \textbf{{Extrinsic parameter $\mathbf{R_c^b}$ estimation.}}
$\mathbf{R}_c^b$ and $\mathbf{q}_c^b$ denote the same rotation from the camera coordinate system to the IMU system as the rotation matrix and the quaternion, respectively. We assume that the relative rotation between the camera and IMU remains constant, and from $i$-th and $i+1$-th frame, we can obtain Eq.~(\ref{eq:k}):
\begin{equation}
\label{eq:k}
\mathbf{q}_b^c\otimes \mathbf{q}_{b_{i+1}}^{b_{i}}=\mathbf{q}_{c_{i+1}}^{c_{i}}\otimes{\mathbf{q}_b^c},
\end{equation}
{where $\otimes$ is quaternion multiplication,} $\mathbf{q}^{b_i}_{b{i+1}}$ is computed using IMU integration, and $\mathbf{q}^{c_i}_{c{i+1}}$ is computed using the epipolar constraint and essential matrix decomposition.
Transforming the above Eq.~(\ref{eq:k}) using quaternion matrices rules, we can obtain the equation as shown in~Eq.~(\ref{eq:2}):
\begin{equation}
\label{eq:2}
\left\{ \left[ \mathbf{q}_{b_{i+1}}^{b_{i}} \right]_L-\left[ \mathbf{q}_{c_{i+1}}^{c_{i}} \right]_R \right\}\mathbf{q}_b^c=0.
\end{equation}
Using the multi-frame information, we can estimate $\mathbf{q}_b^c$ via Singular Value Decomposition (SVD), and then obtain $\mathbf{R}_c^b$ using the commonly-used quaternion-to-rotation matrix transformation algorithm.

\noindent\textbf{Vision-only SfM.}
We start by identifying two frames with a large degree of parallax.
We then use the epipolar constraint to calculate the essential matrix.
By decomposing this matrix, we obtain four possible sets of solutions.
To select the correct solution, we use a method that chooses {$\mathbf{R}^{c_j}_{c_i}$ and $\mathbf{t}^{c_j}_{c_i}$} that maximize the number of feature point pairs while satisfying the condition that the dot product between the landmarks and the feature point is always positive, as shown in Eq.~(\ref{eq:landmarks}) and in Fig.~\ref{fig:E}:
\begin{equation}
\label{eq:landmarks}
    \overrightarrow{O_1P_1}\cdot \overrightarrow{\mathbf{x_1}}>0\ \&\&\ \overrightarrow{O_2P_1}\cdot \overrightarrow{\mathbf{x_2}}>0.
\end{equation}
Using the obtained $\mathbf{R}$ and $\mathbf{t}$, we can triangulate the landmarks $\begin{bmatrix} x_l,y_l,z_l\end{bmatrix}^T$ via Eq.~(\ref{eq:triangulate}):
\begin{equation}
\label{eq:triangulate}
\lambda\begin{bmatrix} \alpha\\\beta\\\gamma\end{bmatrix}={\mathbf{R}_{c_i}^w}\begin{bmatrix} x_l\\y_l\\z_l \end{bmatrix}+{\mathbf{t}_{c_i}^w},
\end{equation}
where $\lambda$ is the scale factor, $\begin{bmatrix} \alpha,\beta,\gamma\end{bmatrix}$ is the unit vector representing the landmark, and $\mathbf{R}$ and $\mathbf{t}$ are the rotation matrix and the translation vector respectively, obtained from the previous step.

\begin{figure}[t!]
	\centering
	\includegraphics[width=1.0\linewidth]{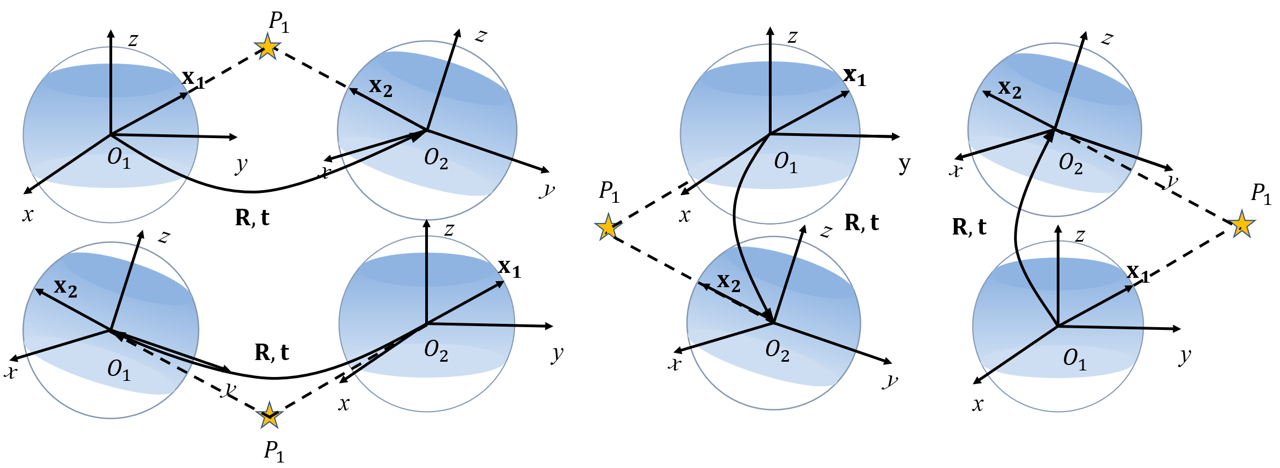}
	\caption{Essential matrix decomposition. There are four sets of solutions to the essential matrix decomposition, but only the first set of solutions is in line with the actual situation.}
	\label{fig:E}
\end{figure}

We use the EPnP method~\cite{lepetit2009epnp} on the landmarks to obtain the {$\mathbf{R}_{c_i}^w$} and {$\mathbf{t}_{c_i}^w$} between the initialization frames. We then alternate between using triangulation and EPnP methods to acquire the three-dimensional spatial positions of all {$\mathbf{R}_{c_i}^w$, $\mathbf{t}_{c_i}^w$,} and landmarks in the sliding window. Finally, we resolve {$\mathbf{R}_{c_i}^w$ and $\mathbf{t}_{c_i}^w$,} using the re-projection error to construct the optimization problem as shown in Eq.~(\ref{eq:3}):
\begin{equation}
\label{eq:3}
{\underset{\mathbf{R}_{c_i}^{w},\mathbf{t}_{c_i}^{w}}{min}\left\{ \sum_{l=1}^M\sum_{m=1}^N||\begin{bmatrix} \alpha_l\\\beta_l\\\gamma_l\end{bmatrix}-\frac{ \mathbf{R}_{c_j}^{c_i}\mathbf{p}_m+\mathbf{t}_{c_j}^{c_i}}{||\mathbf{R}_{c_j}^{c_i} \mathbf{p}_m+\mathbf{t}_{c_j}^{c_i}||_2}||^2\right\},}
\end{equation}
where $M$ is the number of vectors transformed by image observations and $N$ is the number of landmarks.

\noindent\textbf{Visual-inertial alignment.}
The process of aligning visual and inertial data is illustrated in Fig.~\ref{fig:align}.
We match the visual structure obtained from SfM with the IMU pre-integration data.
We then calibrate the gyroscope bias by solving the optimization problem as shown in Eq.~(\ref{eq:4})(\ref{eq:gamma}):
\begin{equation}
\label{eq:4}
\min_{\mathbf{\delta b}_w}{\sum_{k\in all\_frame}{||{\mathbf{q}^{c_{0}}_{b_{k+1}}}^{-1}\otimes \mathbf{q}^{c_{0}}_{b_{k}}\otimes \mathbf{\Gamma}^{b_{k}}_{b_{k+1}}-\begin{bmatrix} 1\\0\\0\\0 \end{bmatrix}||}}^2,\\ 
\end{equation}
\begin{equation}
\label{eq:gamma}
\mathbf{\Gamma}^{b_{k}}_{b_{k+1}}\approx\hat{\mathbf{\Gamma}}^{b_{k}}_{b_{k+1}}\otimes \begin{bmatrix} 1\\ \frac{1}{2}\mathbf{J}_{\mathbf{b}w}^\mathbf{\Gamma}\mathbf{\delta b}_w\end{bmatrix},\\ 
\end{equation}
where $\mathbf{\Gamma}^{b_{k}}_{b_{k+1}}$ and $\hat{\mathbf{\Gamma}}^{b_{k}}_{b_{k+1}}$
are the true-state and nominal-state rotation increment between frame $k$ and frame $k+1$, $\mathbf{\delta b}_w$ is the gyroscope bias, and $\mathbf{J}$ is the Jacobian matrix. Afterward, we estimate the initial velocity, gravity vector, and metric scale, and align the gravity direction with the $Z$ axis.

\begin{figure}[t!]
	\centering
	\includegraphics[width=1.0\linewidth]{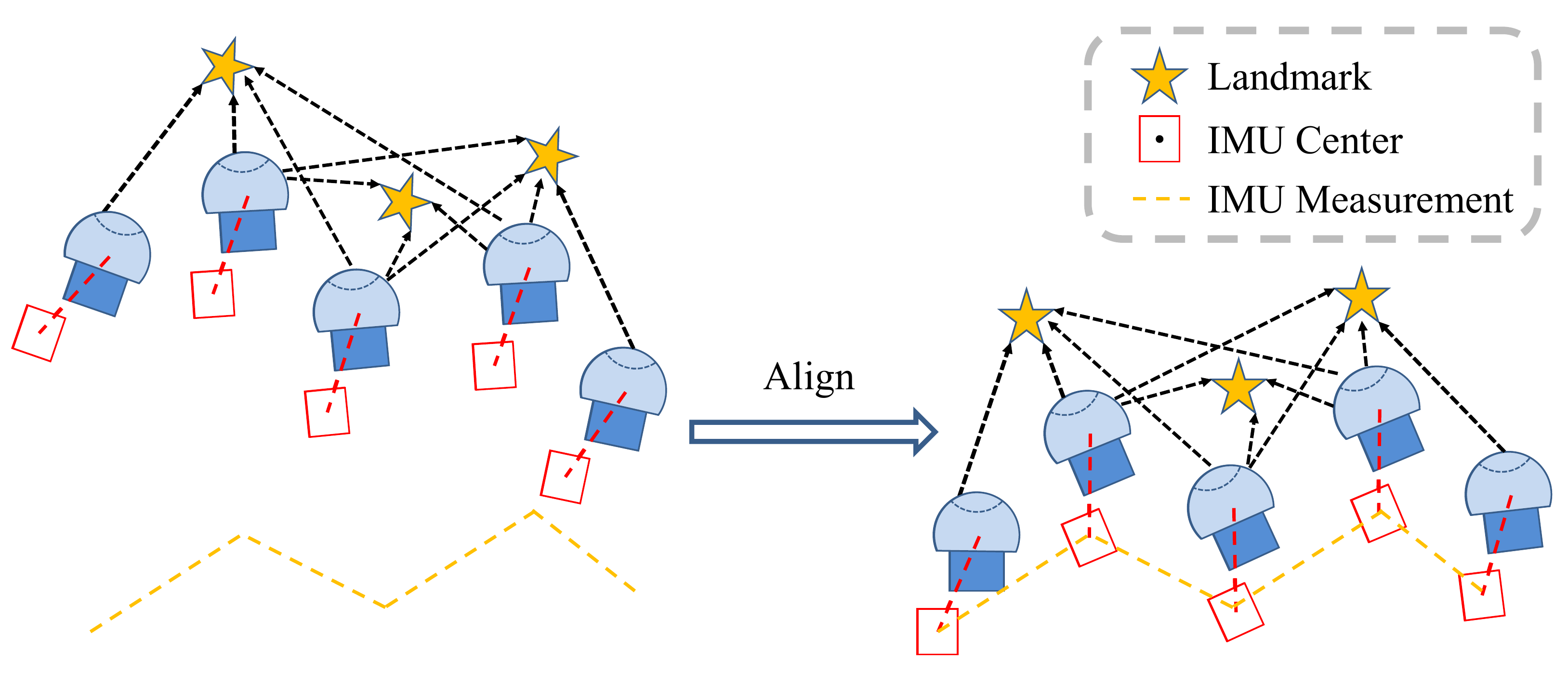}
	\caption{IMU and visual alignment. The up-to-scale visual structure is aligned with the pre-integrated IMU measurements.
	}
	\label{fig:align}
\end{figure}

Our novel initialization method incorporates a suite of algorithms that address the epipolar constraint, triangulation, and re-projection error while considering the feature vector represented by a unit vector.

\subsection{Tightly-coupled Large-Fov Monocular Odometry with a Sliding Window}\label{sec:Odometry}

After the estimation process, we employ a sliding-window-based tightly-coupled monocular odometry for state estimation. {The state vector $\mathbf{\chi}$ is defined as in Eq.~(\ref{eq:state_vector}):}
   \begin{equation}
    \label{eq:state_vector}
    \begin{aligned}
    \mathbf{\chi} & =[\mathbf{s}_1,\dots,\mathbf{s}_{N},{{\mathbf{o}_{c^0}^b,\dots,\mathbf{o}_{c^N}^b}},\lambda_{d_1},\dots,\lambda_{d_m}],\\
     \mathbf{s}_k & =[{\mathbf{p}_{b_k}^w,\mathbf{v}_{b_k}^w,\mathbf{q}_{b_k}^w,} \mathbf{b}_a,\mathbf{b}_g],\\
    {\mathbf{o}_{c^i}^b} &= [{\mathbf{p}_{c^i}^b,\mathbf{q}_{c^i}^b}].
    \end{aligned}
    \end{equation}
Here, $N$ represents the total number of sliding windows, {$\mathbf{p}_{b_k}^w$, $\mathbf{v}_{b_k}^w$, and $\mathbf{q}_{b_k}^w$ are the position, velocity, and attitude quaternion of the body system in the world system in the $k$th frame,} $\mathbf{b}_a$ and $\mathbf{b}_g$ are the accelerometer and gyroscope biases, {$\mathbf{p}_{c^i}^b$ and $\mathbf{q}_{c^i}^b$ are the translation and attitude quaternion of the camera system in the body system in the $i$th frame}. The inverse distance of the $m$th feature from its first observation to the unit sphere is represented by $\lambda_{d_m}$.

During the optimization process, the calibration parameters $\mathbf{t}_{c^N}^B$ will be updated and converge to a reasonable value. The optimization takes into account IMU measurements, visual observations, and marginalization residuals. The optimization objective is shown in Eq.~(\ref{eq:5}):
\begin{equation}
\label{eq:5}
\begin{aligned}
\min\limits_{\chi}
\Bigg\{
||\mathbf{r}_p-\mathbf{H}_p\chi||^2
+&\sum_{k\in\mathcal{B}}||{\mathbf{r}_\mathcal{B}\left(\hat{\mathbf{z}}_{b_{k+1}}^{b_k},\chi\right)||^2_{\mathbf{P}^{b_k}_{b_{k+1}}}}\\
+&\sum_{\left(i,j\right)\in\mathbf{C}}||\mathbf{r}_\mathbf{C}\left(\hat{\mathbf{z}}_{l}^{c_j},\chi\right)||^2_{\mathbf{P}^{c_j}_{l}}
\Bigg\},
\end{aligned}
\end{equation}
where $\mathbf{r}_\mathcal{B}\left(\hat{\mathbf{z}}_{b_{k+1}}^{b_k},\chi\right)$ and $\mathbf{r}_\mathbf{C}\left(\hat{\mathbf{z}}_{l}^{c_j},\chi\right)$ are the IMU measurement residual~\cite{forster2016manifold} and the camera measurement residual, respectively. $\textbf{r}_p$ and $\textbf{H}_p$ represent prior information. The optimization is performed within a sliding window to maintain a low computational complexity.

\noindent\textbf{Visual measurement residual.}
To handle the negative plane of large-FoV panoramic cameras, we propose using a unit sphere to define our visual residual. The $\lambda_d$ represents the inverse distance from the unit sphere to the $l$th feature point observed in the $i$th image. Our visual measurement residual is defined as in Eq.~(\ref{eq:visual_measurement_residual}):
\begin{equation}
\label{eq:visual_measurement_residual}
\textbf{r}_c\left( \hat{\mathbf{z}}_l^{c_j},\chi \right)=\left[
    \begin{array}{cc}
\textbf{b}_1 & \textbf{b}_2
    \end{array}
\right]^T \cdot\left( \hat{\bar{\mathcal{P}}}_l^{c_j}-\frac{\mathcal{P}_l^{c_j}}{||\mathcal{P}_l^{c_j}||}\right),
\end{equation}
\begin{equation}
\begin{aligned}
{\mathcal{P}}_l^{c_j} = & \textbf{R}_b^c\bigg( \textbf{R}_w^{b_j}\bigg( \textbf{R}_{b_i}^w \bigg(  \textbf{R}_{c}^{b} \frac{1}{\lambda_d}\pi^{-1}_s\bigg({\begin{bmatrix}
\hat{\bar{x}}_l^{c_i}\\\hat{\bar{y}}_l^{c_i}
\end{bmatrix}}\bigg)\\
& + {\textbf{p}_{c}^b} \bigg)+{\textbf{p}_{b_i}^w}-{\textbf{p}_{b_j}^w} \bigg)-{\textbf{p}_c^b} \bigg),
\end{aligned}
\end{equation}
where $\textbf{b}_1$ and $\textbf{b}_2$ are two orthogonal bases that span the tangent plane of $\hat{\bar{\mathcal{P}}}_l^{c_j}$, and $\mathcal{P}_l^{c_j}$ is computed using rotation matrices, translation vectors, and the inverse mapping relationship from the unit 3D coordinate point to the pixel point. We use the Ceres Solver~\cite{ceres} to solve the nonlinear maximum posterior estimation problem and a Huber loss function to reduce the influence of outliers for better system robustness.

\subsection{Loop Closure}\label{sec:loop}
Our proposed loop closure system is designed to work in conjunction with our VIO system described above. The loop closure system operates independently of the VIO. It requires input such as the position of feature points in the image, their corresponding 3D coordinates in the camera coordinate system, as well as the current rotation matrix $\mathbf{R}^w_c$ and translation vectors $\mathbf{T}^w_c$ of the VIO system at time $t$. The overall flow of the VIO system with loop closure is illustrated in Fig.~\ref{fig:flow}.
When receiving information from the VIO system, we first apply the pose information to transform the image. To extract meaningful information from the transformed image, we employ the DBoW method~\cite{galvez2012bags}. DBoW converts the extracted feature points' descriptors into a Bag-of-Words (BoW) vector representation. This BoW vector is then used to check for similarities with historical frames in our database.

It is important to note that with large-FoV cameras, loop closure detection can be more prone to errors due to mismatched feature points. To combat this, we propose an improved loop closure detection method that effectively eliminates outliers and includes attitude guidance for enhancing the accuracy and precision of loop closure.

\noindent\textbf{Improved loop closure detection.}
In matching loop-closure frames, feature matching methods are involved and they rely on descriptors, such as BRIEF~\cite{calonder2010brief}, SuperPoint~\cite{detone2018superpoint}, and GMS~\cite{bian2017gms}.
For large-FoV cameras with a negative imaging plane, when the attitude difference is too large, the descriptor information difference is significant.
Therefore, direct matching of the original image is often not very effective, which will be unfolded in Sec.~\ref{sec:Experiments}.
Compared with pinhole cameras, cameras with a large FoV have a higher probability of matching in the loop closure (see a comparison of the FoV of a conventional pinhole camera and a panoramic annular camera in Fig.~\ref{fig:fov_benchmark}).

\begin{figure}[t!]
    \centering
    \includegraphics[width=0.48\textwidth]{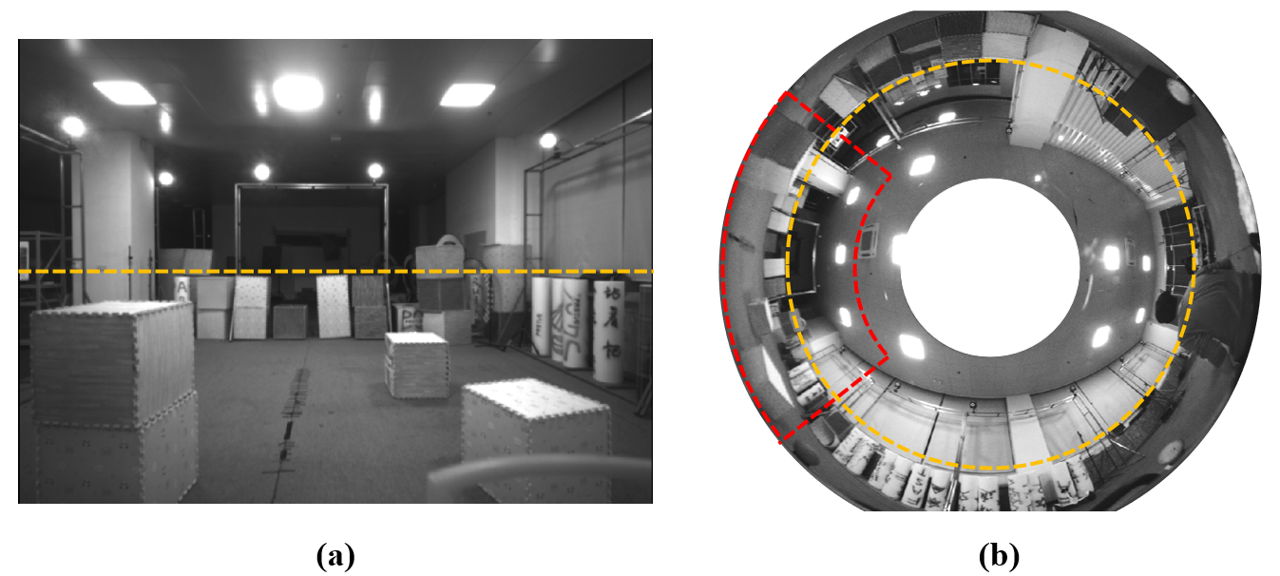}
    \caption{Comparison on FoV. \textbf{(a)} The pinhole camera's FoV is  $87^\circ{\times}58^\circ$. \textbf{(b)} The panoramic camera's FoV is $360^\circ{\times}(40^\circ{\sim}120^\circ)$. The orange dashed line is the horizontal line of the stationary position. The red dashed line denotes the area \textbf{(a)} in the FoV of \textbf{(b)}.}
    \label{fig:fov_benchmark}
\end{figure}

Yet, at the same time, the loop closure detection rate is not high, due to the large difference between the pose of the loop-closure frames and the keyframes.
Therefore, we propose to use the pose information provided by VIO to remap the images and then extract the descriptors, which can greatly improve the success rate of descriptor matching.
For each point in the original image denoted as $[u,v]^T$, the coordinates of the transformed point are $[u_1,v_1]^T$, and the transformation relationship is shown in Eq.~(\ref{equ:transformation}).
\begin{equation}
\label{equ:transformation}
    \begin{bmatrix}u_1\\v_1\end{bmatrix}= \pi_s\left(\mathbf{R}^w_b \mathbf{R}^b_c\pi^{-1}_s\left( \begin{bmatrix} u\\v\end{bmatrix}\right)\right),
\end{equation}
where $\mathbf{R}^w_b$ is the rotation matrix from the body coordinate system to the world coordinate system, $\mathbf{R}^b_c$ is the rotation matrix from the camera coordinate system to the body coordinate system, and $\pi_s$ means the mapping relationship from the unit 3D coordinate point to the pixel point.

\noindent\textbf{Epipolar constraint RANSAC.}
Since features have more outlier points in loop-closure matching than in a conventional VIO system (\textit{i.e.}, in the neighboring-frame matching), if the feature points are incorrectly matched, it will seriously affect the pose solving between loop-closure frames and keyframes, thus affecting the accuracy of loop-closure optimization, so the RANSAC method is crucial in loop-closure threads.
In loop closure, the pair of two points $\mathbf{x}_1$, $\mathbf{x}_2$, is obtained via descriptors matching as shown in Fig.~\ref{fig:ERANSAC}, and they are used for solving the pose via Eq.~(\ref{equ:pose_1})(\ref{equ:pose_2}):
\begin{equation}
\label{equ:pose_1}
{
 {\mathbf{x}_2^T\left( \left[\mathbf{t}^{c_j}_{c_i}\right]^\wedge \mathbf{R}^{c_j}_{c_i} \right)\mathbf{x}_1=0}
},
\end{equation}
\begin{equation}
\label{equ:pose_2}
\mathbf{x}_1=\begin{bmatrix} \alpha_1 \\\beta_1 \\\gamma_1\end{bmatrix},\mathbf{x}_2=\begin{bmatrix} \alpha_2 \\\beta_2 \\\gamma_2\end{bmatrix},
\end{equation}
where the notation $[\mathbf{t}^{c_j}_{c_i}]^{\wedge}$ denotes the skew-symmetric cross product matrix of $\mathbf{t}{\in}{\mathbb{R}^3}$.
Given $\mathbf{x}_1$ , $\mathbf{t}^{c_j}_{c_i}$, and $\mathbf{R}^{c_j}_{c_i}$, if $\mathbf{x}_2$ satisfies the epipolar constraint, ${-}\mathbf{x}_2$ must also satisfy the constraint. This does not happen with cameras that have only a positive imaging half-plane.
{For large-FoV cameras with a negative plane, the number of outlier points, which can be rejected using only the epipolar constraint, is relatively small. The point pairs matched via the descriptor in the loop closure thread have a higher error rate compared to the point pairs matched via the optical flow in the VIO system during initialization.}
Therefore, we propose a novel method to avoid this situation.
Points on the red epipolar line, as depicted in Fig.~\ref{fig:ERANSAC}, can be effectively eliminated in our method.
Our epipolar constraint RANSAC method has the following four steps:
\begin{itemize}
	\item[1)] Each time $8$ points are selected from all pairs of points, and then the essence matrix is calculated and the score of each point is calculated. We eliminate outlier points that do not satisfy the epipolar constraint.
	\item[2)] The decomposition of the highest scoring essential matrix yields four groups $\mathbf{R}_m$ and $\mathbf{t}_m (m{=}1,2,3,4)$.
	\item[3)] The dot product between $\mathbf{Q}_i$ and $\mathbf{R}_m\mathbf{P}_i$ should be greater than the threshold, and again we use the scoring mechanism to eliminate the remaining outlier points.
\end{itemize}
This process is summarized via pseudocode in \textbf{Algorithm}~\ref{alg:E_RANSAC}.

\begin{figure}[!t]
	\centering
	\includegraphics[width=1.0\linewidth]{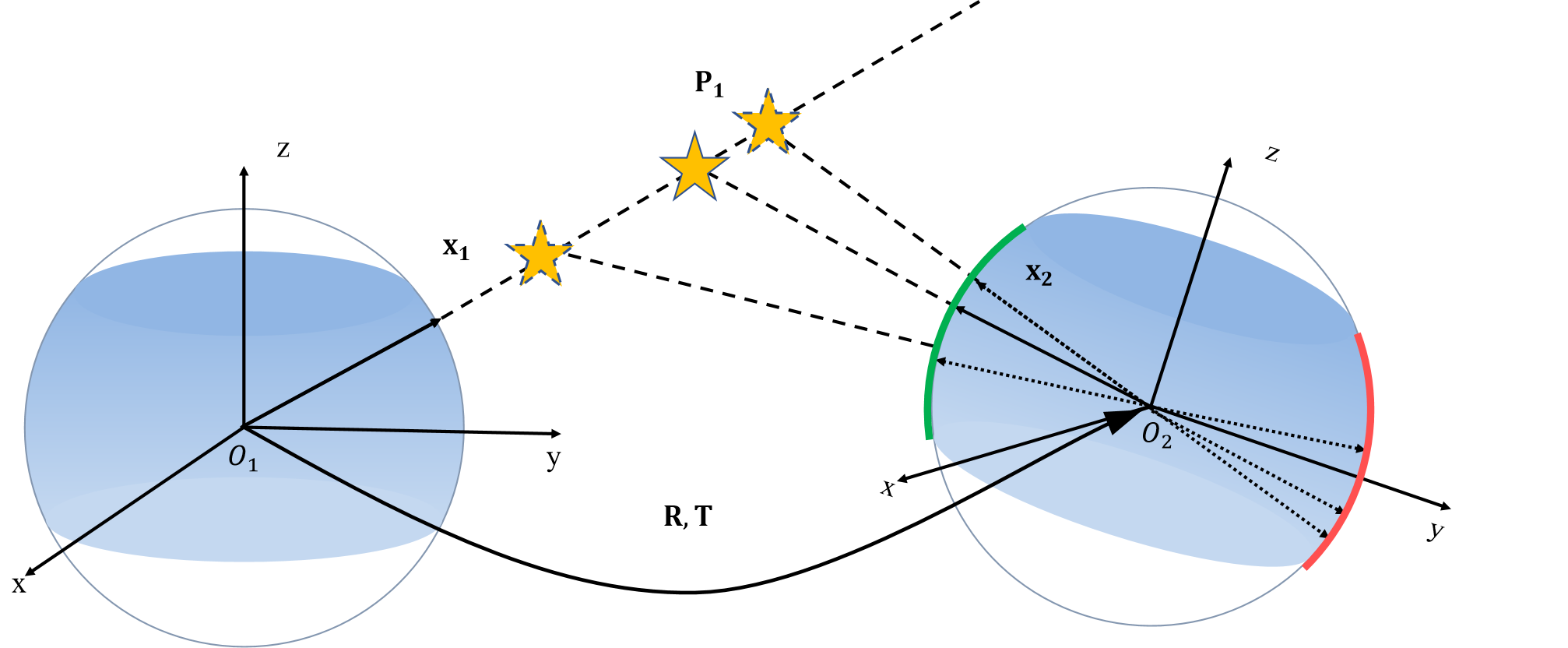}
	\caption{Epipolar constraint. The \orange{orange} pentagrams indicate landmarks and each pentagram corresponds to the two vectors on the right, where vector vertices on the \green{green} line are retained and on the \red{red} line are removed.}
	\label{fig:ERANSAC}
\end{figure}

\begin{algorithm}[t]
\small
	\caption{Epipolar Constraint RANSAC Method}
	\label{alg:E_RANSAC}
	\KwIn{pairs of feature points sets $\{\mathbf{P}\}$, $\{\mathbf{Q}\}$, iteration number $K$}
	\KwOut{points status sets $\{S\}$ (1: inlier point, 0: outlier point), the flag that the algorithm has a solution $flag$ (1: success, 0: failure)}
	\Begin
	{
	    Initialization: $\mathbf{E}_{best}$,$best\_score\leftarrow 0$
	    $match\_num\leftarrow$elementNumber($\{P\}$)\;
	    \If{$match\_num<8$}
	    {$flag\leftarrow 0$\;
	    \Return $flag$\;}
        \For{$i\leftarrow 1$ \KwTo $K$}
        {
          Randomly choose 8 pairs of points $\{\mathbf{P}_j\}$, $\{\mathbf{Q}_j\}$\;
          $\mathbf{E}\leftarrow$ComputeEssentialMatrix($\{\mathbf{P}_j\}$, $\{\mathbf{Q}_j\}$)\;
          $score,\{S_{tmp}\}\leftarrow$CheckInliers($\mathbf{E}$, $\{\mathbf{P}\}$, $\{\mathbf{Q}\}$)\;
          \If{$score>best\_score$}
	      {
	      $best\_score\leftarrow score$\;
	      $\mathbf{E}_{best}\leftarrow \mathbf{E}$\;
	      $\{S_{best}\}\leftarrow\{S_{tmp}\}$\;
	      }
        }
        $\mathbf{R}_m,\mathbf{t}_m\leftarrow$DecomposeEssentialMat($\mathbf{E}_{best}$)\;
        $\{\mathbf{C}_i\}\leftarrow$triangulatePoints($\mathbf{R}_m$,$\mathbf{t}_m$,$\{\mathbf{P}_i\}$, $\{\mathbf{Q}_i\}$)\;
        \For{$i\leftarrow 1$ \KwTo $match\_num$}
        {
            \For{$j\leftarrow 1$ \KwTo $m$}
           { 
                    $PdotC=\mathbf{P}_i^T\mathbf{C}_i/||\mathbf{C}_i||_2$\;
                    $\mathbf{C_{tmp}}=\mathbf{R}_j\mathbf{C}_i+\mathbf{t}_j$\;
                    $QdotC=\mathbf{Q}_i^T\mathbf{C_{tmp}}/||\mathbf{C_{tmp}}||_2$\;
                    \If{$PdotC<th$ and $QdotC<th$ and $\mathbf{Q}_i^T\mathbf{R}_j \mathbf{P}_i<th2$}
                    {
                        $good_{j}\leftarrow good_{j}+1$\;
                        $S_j[i]\leftarrow 0$;
                    }
            }
        }
        \If{$good_{a}>=good_{b}$}
        {
            \For{$i\leftarrow 1$ \KwTo $match\_num$}
            {
                \If{$S_a[i]==0$}
                {
                    $S_{best}[i]\leftarrow0$
                }
            }
        }
        $flag\leftarrow 1$,$\{S\}\leftarrow\{S_{best}\}$\;
		\Return{$flag$, $\{S\}$}\;
	}
\end{algorithm}

\subsection{EPnP RANSAC for Unit Vector Feature Points}
\label{sec:epnp_ransac_unit_vector_feature_points}
Since we use the unit vectors as introduced to characterize the feature points, EPnP~\cite{lepetit2009epnp} needs to consider the negative imaging plane case when checking the inlier points.
We make sure that the angle between the map point and the feature point in the camera coordinate system is less than the threshold value, to filter out inlier points.
The pseudocode of this process is summarized in \textbf{Algorithm}~\ref{alg:pnp_RANSAC}.
\begin{algorithm}
\small
	\caption{Our EPnP RANSAC Method}
	\label{alg:pnp_RANSAC}
	\KwIn{pairs of feature points set and matched landmark Points $\{\mathbf{P}\}$, $\{\mathbf{W}\}$, iteration number $K$}
	\KwOut{Rotation Matrix $\mathbf{R}$, translation vector $\mathbf{t}$,
	points status sets $\{S\}$,the flag that the algorithm has a solution $flag$}
	\Begin
	{
	    Initialization: $best\_score\leftarrow 0$
	    $match\_num\leftarrow$elementNumber($\{\mathbf{P}\}$)\;
	    \If{$match\_num<4$}
	    {$flag\leftarrow 0$\;
	    \Return $flag$\;}
        \For{$i\leftarrow 1$ \KwTo $K$}
        {
          Randomly choose 4 pairs of points $\{\mathbf{P}_j\}$, $\{\mathbf{W}_j\}$\;
          $\mathbf{R}_{tmp},\mathbf{t}_{tmp}\leftarrow$ComputePoseByEpnp($\{\mathbf{P}_j\}$, $\{\mathbf{W}_j\}$)\;
          $score,\{S_{tmp}\}\leftarrow$CheckInliers($\mathbf{R}_{tmp}$, $\mathbf{t}_{tmp}$, $\{\mathbf{P}\}$, $\{\mathbf{W}\}$)\;
          \If{$score>best\_score$}
	      {
	      $best\_score\leftarrow score$\;
	      $\mathbf{R}\leftarrow \mathbf{R}_{tmp}$\;
	      $\mathbf{t}\leftarrow \mathbf{t}_{tmp}$\;
	      $\{S_{best}\}\leftarrow\{S_{tmp}\} $\;
	      }
        }
        $flag\leftarrow 1$,$\{S\}\leftarrow\{S_{best}\}$\;
		\Return{$\mathbf{R}$, $\mathbf{t}$, $flag$, $\{S\}$}\;
	}
\end{algorithm}

\noindent\textbf{Pose graph optimization.}
\label{pose_graph_optimization}
Pose graph optimization plays a role in error correction in loop closure thread.
To optimize each node in the pose graph, we minimize the following cost function to optimize the whole graph of loop-closure edges and sequential edges via Eq.~(\ref{equ:pose_graph_1})(\ref{equ:pose_graph_2}):
\begin{equation}
\label{equ:pose_graph_1}
    \min\limits_{\mathbf{p},\psi}\left\{ \sum\limits_{\left( i,j\in\mathcal{L}\right) }||{\mathbf{r}_{i,j}||^2}+ \sum\limits_{\left( i,j\in\mathcal{S}\right) }||\rho\left(\mathbf{r}_{i,j}\right)||^2\right\},
\end{equation}
\begin{equation}
\label{equ:pose_graph_2}
    \mathbf{r}_{i,j}\left( \mathbf{p}^w_i,\psi_i,\mathbf{p}^w_j,\psi_j \right)=\begin{bmatrix}\mathbf{R}\left( \hat{\phi}_i,\hat{\theta}_i,\psi_i \right)^{-1}\left( \mathbf{p}^w_j-\mathbf{p}^w_i \right)-\hat{\mathbf{p}^i_{ij}}\\\psi_j-\psi_i-\hat{\psi_{ij}}\end{bmatrix},
\end{equation}
where $\hat{\phi}_i$ and $\hat{\theta}_i$ are the fixed estimates of roll and pitch angles because, for a VIO system, the accelerometer prevents divergence of roll and pitch angles. $\rho$ is the Huber norm function. $\mathcal{S}$ and $\mathcal{L}$ are the sets of all sequential edges and loop-closure edges. 

Pose graph saving and loading allow us to use priori maps for loop closure. We only save descriptors of every keyframes and the $i$th keyframe's state as shown in Eq.~(\ref{equ:state}):
\begin{equation}
\label{equ:state}
\begin{bmatrix}i,\hat{\mathbf{p}^w_i}, \hat{\mathbf{q}^w_i},v,\hat{\mathbf{p}^i_{iv}},\hat{\psi_{iv}},D\left( \alpha,\beta,\gamma,\mathbf{des} \right)\end{bmatrix},
\end{equation}
where $i$, $\hat{\mathbf{p}^w_i}$, and $\hat{\mathbf{q}^w_i}$ are the frame index, position, and orientation. $v$, $\mathbf{p}^i_{iv}$, and $\hat{\psi_{iv}}$ are the loop closure index,
relative position, and yaw. $D\left( \alpha,\beta,\gamma,\mathbf{des} \right)$ is the feature set containing the unit 3-D location and its descriptor with our attitude-guided transformation.

\section{Experiments}\label{sec:Experiments}
In this section, we conduct experiments to confirm the effectiveness of the proposed components and the LF-VISLAM system.
In Sec.~\ref{exp:datasets}, we describe the datasets used in our experiments.
In Sec.~\ref{exp:fov}, we study the effects using different FoV inputs for the performance of visual inertial odometry and certify the importance of using the information on the negative imaging plane.
In Sec.~\ref{exp:lidar}, we verify our method is beneficial when integrated with LiDAR-visual-inertial odometry.
We further evaluate our attitude-guided descriptor extraction algorithm in Sec.~\ref{exp:attitude_guided_descriptors}.
We compare our LF-VISLAM system against state-of-the-art SLAM methods in Sec.~\ref{exp:comparison}, specifically analyze the results on long sequences in Sec.~\ref{exp:long}, and finally assess the efficiency of our system in Sec.~\ref{exp:speed}.

\begin{table*}[!t]
 \setlength{\tabcolsep}{2.2pt}
	\centering
	\begin{threeparttable}
	\caption{Accuracy analysis of LF-VIO using images with different FoVs on the PALVIO ID06 set.}
	\label{tab:different_fov}
\renewcommand\arraystretch{1.4}{\setlength{\tabcolsep}{7.5mm}{\begin{tabular}{cccccc}
\toprule
Field of View & $40^{\circ}{\sim}120^{\circ}$ & $40^{\circ}{\sim}110^{\circ}$ & $40^{\circ}{\sim}100^{\circ}$ & $40^{\circ}{\sim}90^{\circ}$ & $40^{\circ}{\sim}80^{\circ}$ \\
\midrule
\midrule
RPEt (\%) &\textbf{2.814} & 3.097 & \underline{2.923} & 2.933 & 3.392 \\
RPEr (degree/m) & \underline{0.397} & \textbf{0.393} & 0.402 & 0.433 & 0.631 \\
ATE (m) &\textbf{0.093}& 0.182 & \underline{0.112} & 0.124 & 0.171 \\

\midrule
Field of View & $50^{\circ}{\sim}120^{\circ}$ & $60^{\circ}{\sim}120^{\circ}$ & $70^{\circ}{\sim}120^{\circ}$ & $80^{\circ}{\sim}120^{\circ}$ & $90^{\circ}{\sim}120^{\circ}$ \\
\midrule
RPEt (\%) 
&\textbf{2.585}&\underline{2.712}&3.584&3.525&5.223
\\
RPEr (degree/m) 
&\textbf{0.394}&\underline{0.398}&0.417&0.449&0.490
\\
ATE (m) 
&\textbf{0.081}&\underline{0.117}&0.250&0.252&0.475
\\
	\bottomrule
	\end{tabular}}}
	\end{threeparttable}
\end{table*}

\subsection{Datasets}
\label{exp:datasets}
\noindent\textbf{PALVIO dataset.}
To address the lack of panoramic visual odometry datasets with ground-truth location and pose, we have created and made publicly available the PALVIO dataset. {This dataset is collected using two panoramic annular cameras, a CUAV-v5 nano IMU sensor, and a RealSense D435 sensor, all of which are synchronized with ground-truth location and pose data captured by a motion capture system.} The panoramic cameras capture monocular images with a resolution of $1280{\times}960$ at a rate of $30Hz$ and a field of view of $360^\circ{\times}(40^\circ{\sim}120^\circ)$. {We record all datasets using Robot Operating System (ROS) and the data are raw without additional processing.} The IMU sensor provides angular velocity and acceleration data at $200Hz$. {The motion capture system (Vicon T40s) provide position and attitude data at $100Hz$, serving as the ground truth.} A total of ten sequences (ID01${\sim}$ID10) are collected in an indoor area of $8m{\times}10m$. The data from both the stereo panoramic cameras and the RealSense camera are available for public use, and in this work, we primarily use the data from the top panoramic camera for our experiments. The displacement of the ID01${\sim}$ID10 dataset is not significant.

To test the robustness and loop-closure effects of our LF-VISLAM, we have also collected long-trajectory sequences IDL01 and IDL02. {The long-trajectory sequences are captured using a mynteye camera module, which includes a panoramic camera, an IMU sensor, and an Ouster LiDAR sensor. The panoramic cameras capture monocular images with a resolution of $1280{\times}720$ at a rate of $30Hz$, the LiDAR sensor provides LiDAR points at $10Hz$, and the IMU sensor provides angular velocity and acceleration data at $200Hz$.}

\noindent\textbf{LVI-SAM dataset.}
To evaluate the generalizability of our approach, we further use the LVI-SAM dataset~\cite{shan2021lvi}. The data collection sensor suite consisted of a Velodyne VLP-16 LiDAR sensor, a FLIR BFS-U3-04S2M-CS camera, a MicroStrain 3DM-GX5-25 IMU sensor, and a Reach RS+ GPS. The $\emph{Jackal}$- and $\emph{Handheld}$ datasets are gathered using an unmanned ground vehicle. The field of view of the fisheye camera used in the experiments is approximately $360^\circ{\times}(0^\circ{\sim}93.5^\circ)$. To evaluate the performance, we use the LVI-SAM dataset, in conjunction with GPS measurements as the ground truth.

\begin{figure}[t]
	\centering
	\includegraphics[width=1.0\linewidth]{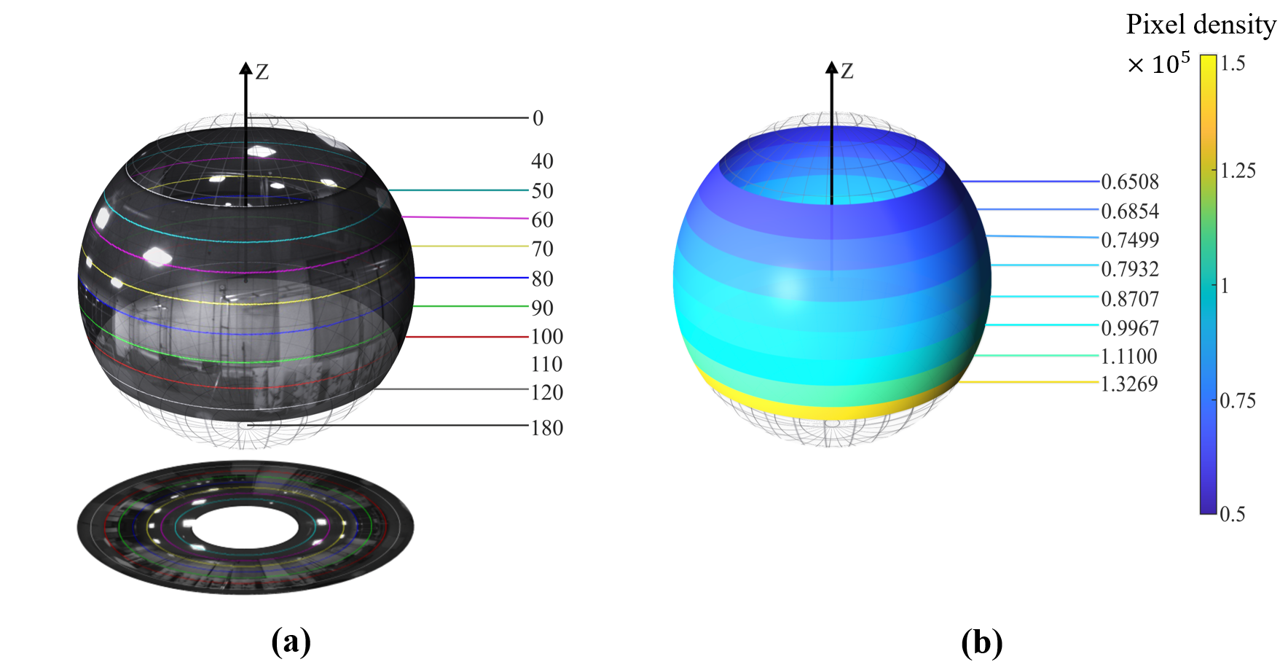}
	\caption{{Variation of \textbf{(a)} FoVs and \textbf{(b)} pixel density with a step of $10^\circ$ using the annular images.}}
	\label{fig:different_fov}
\end{figure}

\begin{table*}[t]
 \setlength{\tabcolsep}{2.2pt}
	\centering
	\begin{threeparttable}
	\caption{Pixel density analysis at different view bands.}
	\label{tab:pixel_density}
\renewcommand\arraystretch{1.4}{\setlength{\tabcolsep}{2.6mm}{\begin{tabular}{ccccccccc}
\toprule 
View band &$40^{\circ}{\sim}50^{\circ}$ & $50^{\circ}{\sim}60^{\circ}$ & $60^{\circ}{\sim}70^{\circ}$ & $70^{\circ}{\sim}80^{\circ}$ & $80^{\circ}{\sim}90^{\circ}$ & $90^{\circ}{\sim}100^{\circ}$ & $100^{\circ}{\sim}110^{\circ}$ & $110^{\circ}{\sim}120^{\circ}$\\
\midrule
\midrule
Pixel of image ($10^5$) &0.50&0.61
&0.74&0.84&0.95&1.08&1.17&	1.31\\
\midrule
\begin{tabular}[c]{@{}c@{}}Annulus area \\ per unit sphere\end{tabular} &0.77&0.89&0.99&1.05&1.09&1.09&1.05&0.99     \\
\midrule
Pixel density ($10^5$) &0.65&0.69&0.75&0.79&0.87&1.00&1.11&1.32  \\
\bottomrule
	\end{tabular}}}
	\end{threeparttable}
\end{table*}

\begin{table*}[t]
 \setlength{\tabcolsep}{2.2pt}
	\centering
	\begin{threeparttable}
	\caption{Comparison of SLAM methods on the LVI-SAM dataset~\cite{shan2021lvi} (without loop).}
	\label{tab:b_lf_lvi_sam}
	\renewcommand\arraystretch{1.4}{\setlength{\tabcolsep}{0.95mm}{\begin{tabular}{cccccccccccccc}
    \toprule    
    \multirow{2}{*}{Dataset}  & \multirow{2}{*}{LVIO-Method}      & \multicolumn{4}{c}{RPEt(\%)} & \multicolumn{4}{c}{RPEr (degree/m)} & \multicolumn{4}{c}{ATE(m)} \\
                          &                                  & Mean   & Min  & Max  & RMSE  & Mean    & Min    & Max    & RMSE    & Mean  & Min  & Max  & RMSE \\ 
    \midrule
    \midrule
    \multirow{2}{*}{$\emph{Handheld}$} & LF-LVI-SAM (Ours) 	& \textbf{0.63857} &  \textbf{0.00660}&4.38248&\textbf{0.84366}&
    \textbf{0.00393}&\textbf{0.00007}&0.02180& \textbf{0.00510}&
    \textbf{8.89108}&\textbf{0.08698}&\textbf{25.30987}&\textbf{11.57598}
	\\
    & LVI-SAM~\cite{shan2021lvi}    	&0.67193&0.01113&\textbf{4.37260}&0.87484&
	0.00397&0.00010&\textbf{0.01947}&0.00511&
	10.78413&0.09040&26.72794&13.52003
	\\
    \midrule 
    \multirow{2}{*}{$\emph{Jackal}$} & LF-LVI-SAM (Ours) & \textbf{0.70924} & \textbf{0.00948}&2.83097& \textbf{0.79693}&
	\textbf{0.00265}&\textbf{0.00047}&\textbf{0.00761}&\textbf{0.00283}&
	\textbf{5.80506}&0.80519&\textbf{14.21643}&\textbf{6.28478}
	\\
    & LVI-SAM~\cite{shan2021lvi} 	
    &0.73586&0.01300&\textbf{2.80597}&0.82791&
	0.00274&\textbf{0.00047}&0.00781&0.00291&
	6.02435&\textbf{0.65769}&14.53112&6.53074
	\\  
	
	\bottomrule
	\end{tabular}}}
	\end{threeparttable}
\end{table*}

\subsection{Investigation on Different Field-of-View}
\label{exp:fov}
To validate the significance of the information from the negative plane, we evaluate the impact of using inputs with different FoVs by extracting features only from the corresponding angle range. First, we gradually reduce the FoV from $40^{\circ}{\sim}120^{\circ}$ to $40^{\circ}{\sim}80^{\circ}$ with a step of $10^\circ$, as shown in Fig.~\ref{fig:different_fov}, and conduct experiments using our LF-VIO framework on the PALVIO ID06 dataset, which is selected as a representative sequence.
The results, shown in Table~\ref{tab:different_fov}, indicate that using the entire FoV of the panoramic camera ($40^{\circ}{\sim}120^{\circ}$) results in the best performance in terms of RPEt and ATE.
However, when the FoV is reduced to only the positive plane, \textit{i.e.}, in the cases of $40^{\circ}{\sim}90^{\circ}$ and $40^{\circ}{\sim}80^{\circ}$, the performance significantly degrades, verifying the importance of the information from the negative imaging plane.

Next, we repeat the experiment by decreasing the FoV from $50^{\circ}{\sim}120^{\circ}$ to $90^{\circ}{\sim}120^{\circ}$. In the worst-case scenario, when only the information from the negative plane is used ($90^{\circ}{\sim}120^{\circ}$), the VIO framework still performs effectively. However, when incorporating slightly more information from the positive plane, \textit{i.e.}, in the cases of $80^{\circ}{\sim}120^{\circ}$ and $70^{\circ}{\sim}120^{\circ}$, the performance returns to the same level as that of using only the positive plane. This confirms once again the importance of information from the negative plane and the efficacy of our method in making use of negative-plane features that were ignored in prior works.

We also observe that the performance of the FoV range of $50^\circ{\sim}120^\circ$ is not superior to that of $40^\circ{\sim}120^\circ$, due to the lowest vector density in the range of $40^\circ{\sim}50^\circ$. The information from this band may not significantly improve the performance of the system.
To visualize the density of the pixels in the image, we map each pixel onto a unit sphere and divide it into intervals of $10^\circ$, as shown in Fig.~\ref{fig:different_fov}\textbf{(b)} and Table~\ref{tab:pixel_density}.
Each pixel in the original image can be represented by a vector with a magnitude of $1$. However, when the vectors from various FoVs are projected onto the corresponding FoV on the sphere, their densities vary. We calculate the vector density of different field-of-view regions on the unit sphere and find that the density in the range of $40^\circ{\sim}50^\circ$ is the lowest, resulting in the lowest accuracy among all field-of-view bands. Although this band provides significant information, it has a lower pixel accuracy compared to other regions, and thus, incorporating information from this band may not improve the accuracy of the SLAM system.

\subsection{Generalization to LiDAR-Visual-Inertial Odometry}
\label{exp:lidar}
The proposed method, which supports the negative half-plane, can be seamlessly integrated with LiDAR-Visual-Inertial Systems like LVI-SAM~\cite{shan2021lvi}. The accuracy of the system with and without the proposed method are compared in Table~\ref{tab:b_lf_lvi_sam} using two datasets, namely the $\emph{Handheld}$- and $\emph{Jackal}$ datasets provided by~\cite{shan2021lvi}. The experiments are conducted on a laptop with an R7-5800H processor. The recording camera has a FoV of $360^\circ{\times}(0^\circ{\sim}93.5^\circ)$.

In this experiment, we make modifications to the visual component of the LVI-SAM system~\cite{shan2021lvi}. We utilize a unit vector approach to represent feature points and update the visual algorithms accordingly while maintaining the setup of the original MEI camera model~\cite{mei2007single}. The original LVI-SAM system uses a large mask to address the issues associated with feature points on the negative plane or those close to $180^\circ$. However, our system has successfully overcome this challenge by reducing the size of the mask, leading to improved performance. The results of our modified system referred to as LF-LVI-SAM, show consistent improvements against LVI-SAM~\cite{shan2021lvi}, in mean and root mean square error of relative pose error (RPE) in Mean and RMSE of RPE (\%), RPEr (degree/m) and ATE (m). The RPE results on the Handheld and Jackal datasets are presented in Table~\ref{tab:b_lf_lvi_sam}.
In summary, our method not only enhances the precision of visual-inertial odometry systems but is also effective for LiDAR-VIO systems with a FoV extending to the negative half-plane.

\begin{figure*}[!t]
	\centering
    \includegraphics[width=1.0\linewidth]{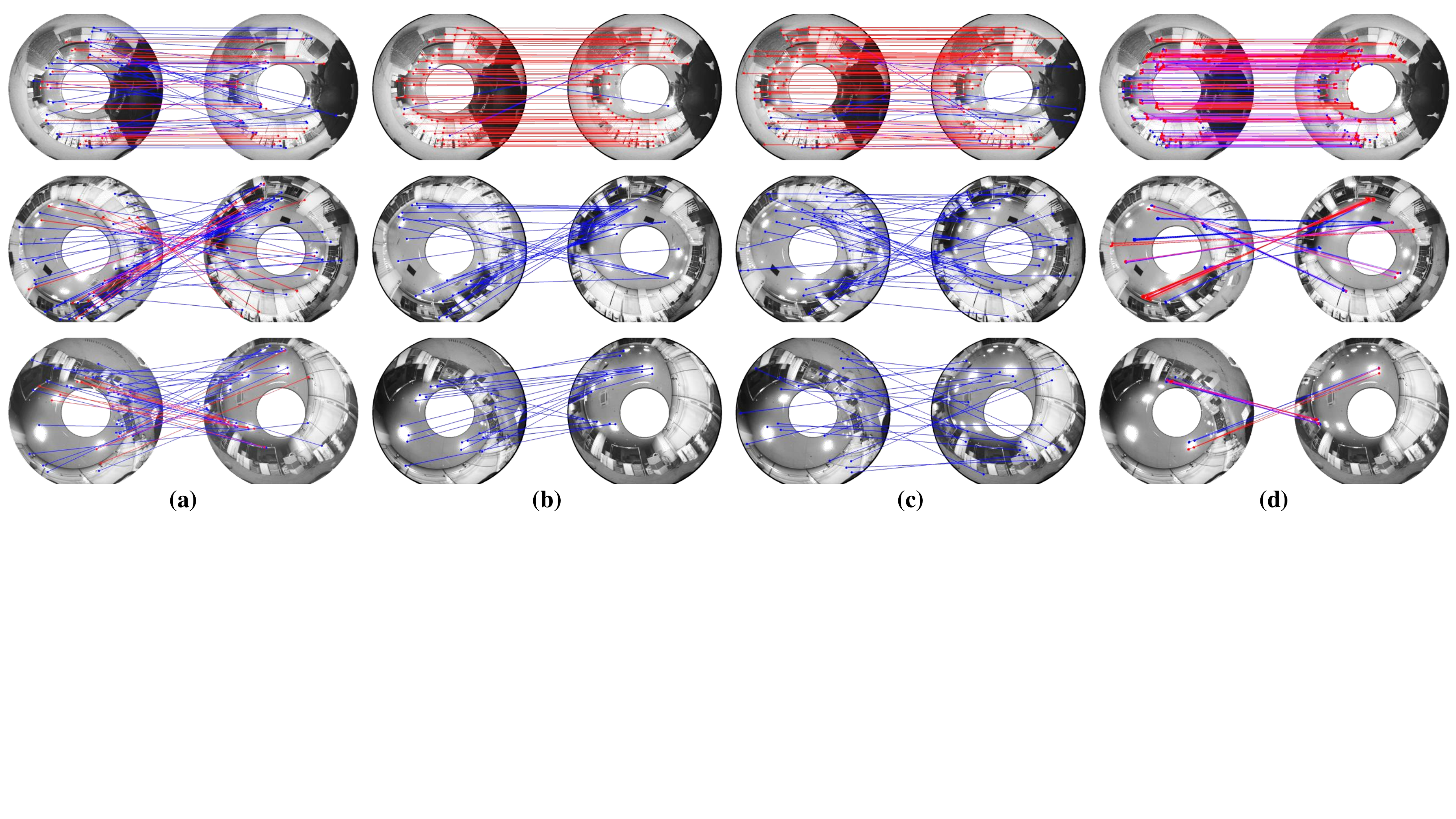}
    \vskip-25ex
	\caption{\textbf{(a)} The BRIEF descriptor with attitude guidance and matching (Ours). \textbf{(b)} The conventional BRIEF descriptor and matching. \textbf{(c)} The SuperPoint descriptor and matching~\cite{detone2018superpoint}. \textbf{(d)} The GMS matching~\cite{bian2017gms}. For illustration purposes, the illumination and contrast of the panoramic images are improved. The three rows indicate scenarios with small angle differences, large yaw angle differences, and large tilt angle differences, respectively. The entire \blue{blue} and red points represent all the matched points, and the \red{red} points are the remaining points after applying our RANSAC method.}
	\label{fig:match}
\end{figure*}

\subsection{On the Effectiveness of Attitude-Guide Descriptors}
\label{exp:attitude_guided_descriptors}
To verify that the introduction of attitude-guided descriptors is more effective than the direct extraction of descriptors, we explore different settings including (a) transforming the images with attitude guidance and extracting BRIEF descriptors for matching using Hamming distance, (b) directly extracting BRIEF descriptors~\cite{calonder2010brief} for matching, (c) directly extracting SuperPoint descriptors~\cite{detone2018superpoint} for matching, and (d) the GMS solution~\cite{bian2017gms} in comparison experiments with small attitude angle differences, large yaw angle differences, and large tilt angle differences, as shown in the three rows in Fig.~\ref{fig:match}, respectively. 
To facilitate a fair comparison, our RANSAC is used to remove the outliers for the above methods.
The original number of ORB points~\cite{campos2021orb} in the GMS is $1000$, and the rest only uses $500$ points.

As shown in the first row, when the difference between the attitude angles of the two images is very small, our method matches more error points compared with the other three methods, but all the outlier points can be removed by the RANSAC method.
As shown in the second row, when the difference between the yaw angles of the two images is large, the conventional BRIEF and the learning-based SuperPoint basically do not match the correct points at all, and the performance of GMS and our method is close.
As shown in the third row, when the yaw angles of the two panoramic images differ significantly, our algorithm has an obvious advantage over the other three algorithms, in which BRIEF and SuperPoint methods struggle to match correct points, and the GMS solution matches relatively more yet all close points, so the effect is rather unsatisfactory. 
Thus, our method is more effective than other algorithms under large attitude angle differences, despite that the learning-based GMS uses a bigger number of original points for matching.
It is possible to perform loop closure even when the angle difference is very large so that the success rate of loop-closure matching can be improved, promising for improving pose estimation accuracy with large-FoV panoramic cameras.

\begin{table*}[!t]
 \setlength{\tabcolsep}{2.2pt}
	\centering
	\begin{threeparttable}
	\caption{Comparison of SLAM methods on the PALVIO dataset. ``w'': with loop closure, ``w/o'': without loop closure.}
	\label{tab:slam}
\renewcommand\arraystretch{1.4}{\setlength{\tabcolsep}{2.7mm}{\begin{tabular}{cccccccccccc}
\toprule   
\multicolumn{2}{c}{\multirow{2}{*}{VIO-Method}} & \multicolumn{10}{c}{Sequences}                                                                                                                                          \\
\multicolumn{2}{c}{}                            & ID01           & ID02           & ID03           & ID04           & ID05           & ID06           & ID07           & ID08           & ID09           & ID10           \\
\midrule
\midrule

\multirow{3}{*}{{LF-VISLAM(w)}}      & {RPEt (\%)}&\textbf{2.861}&\textbf{2.576}&\textbf{2.367}&\textbf{1.379}&\textbf{1.858}& \textbf{2.386}&\textbf{1.647}&\textbf{2.346}&\textbf{1.591}&\textbf{3.525}       \\
                             & {RPEr (degree/m)}&\textbf{0.307}&\textbf{0.362} &\textbf{0.208}&\underline{0.178}&\textbf{0.235}&\underline{0.386}&\underline{0.224}&\textbf{0.192}&\textbf{0.140}&\textbf{0.558}           \\
                             & {ATE (m)} &\textbf{0.229}& \textbf{0.143}& \textbf{0.130} &\textbf{0.129}&\textbf{0.143}&\textbf{0.083}&\textbf{0.138}&\textbf{0.135}&\textbf{0.113}&\textbf{0.180}          \\
\midrule
\multirow{3}{*}{LF-VIO(w/o)~\cite{wang2022lfvio}}      & RPEt (\%)         & \underline{3.556} & \underline{2.709} & \underline{2.542} & \underline{1.495} & \underline{2.016} & {2.814} & \underline{2.775} & 2.983          & 2.146          & \underline{4.493} \\
                             & RPEr (degree/m)  & \underline{0.328} & 0.599 & 0.292          & 0.227          & 0.328          & 0.397          & 0.315          & \underline{0.202} & {0.322} & \underline{0.567}          \\
                             & ATE (m)          & \underline{0.341} & \underline{0.153} & \underline{0.269} & \underline{0.166} & 0.200          & \underline{0.093} & \underline{0.237} & 0.236          & \underline{0.222} & \underline{0.292} \\
\midrule

\multirow{3}{*}{{LF-VIO-40-90(w/o)}}      & {RPEt (\%)}        & {4.245} & {5.707} & {3.468} & {2.571} & {2.361} & {2.933} & {3.051} & {3.050} & {2.954} & {5.255} \\
                             & {RPEr (degree/m)}  & {0.405} & {0.892} & {0.370} & {0.316} & {0.378}          & {0.433}  & {0.368}  & {0.505} & {0.413} & {0.572}          \\
                             & {ATE (m)}           & {0.476} &  {0.583} & {0.401} & {0.312} & {0.327}          & {0.124} & {0.384} & {0.309} & {0.463} & {0.349} \\
\midrule
\multirow{3}{*}{SVO2.0(w)~\cite{forster2014svo}}      & RPEt (\%)         & 6.531          & 6.995          & 2.710          & 1.928          & 2.354          & 3.409          & 3.718          & 2.811          & \underline{2.012}& 14.147          \\
                             & RPEr (degree/m)  & 0.401          & \underline{0.378}         & \underline{0.235} & \textbf{0.165}          & 0.296          & \textbf{0.320} & \textbf{0.187} & 0.291          & \underline{0.230}          & 0.608 \\
                             & ATE (m)           & 0.761          & 0.380          & 0.366          & 0.174          & \underline{0.148} & 0.124          & 0.428          & 0.236          & 0.292          & 1.122          \\
\midrule
\multirow{3}{*}{VINS-Mono(w)~\cite{qin2018vins}}   & RPEt (\%)         & 5.446          & 3.542          & 2.767          & 2.189          & 2.553          & 2.993          & 2.941          & \underline{2.405} & 2.933          & 4.494         \\
                             & RPEr (degree/m)  & 0.458         & 0.605          & 0.285          & 0.249          & \underline{0.278} & 0.445          & 0.339          & 0.452          & 0.457          & \underline{0.567}         \\
                             & ATE (m)           & 0.870          & 0.214          & 0.310          & 0.217          & 0.263          & 0.104          & 0.299          & {0.194} & 0.378          & 0.557   \\ 

	\bottomrule
	\end{tabular}}}
	\end{threeparttable}
\end{table*}

\subsection{Comparison on the Established PALVIO Dataset}
\label{exp:comparison}
Since most of the data in the established PALVIO dataset are collected in indoor areas, these sequences are suitable for loop-closure accuracy experiments.
Our LF-VISLAM is an addition of loop closure threads to LF-VIO~\cite{wang2022lfvio}.
At the same time, we also compare with SVO2.0~\cite{forster2014svo} and VINS-Mono~\cite{qin2018vins}, both of which have loop closure threads for a fair comparison in Table~\ref{tab:slam} and Fig.~\ref{fig:lfvio-vins-svo}.
{We utilize the Relative Pose Error in translation (RPEt)~\cite{Zhang18iros}, Relative Pose Error in rotation (RPEr)~\cite{Zhang18iros}, and Absolute Trajectory Error (ATE)~\cite{Zhang18iros} as the evaluation metrics of the overall system error.
The above four methods all use the Scaramuzza~\textit{et al.}'s omnidirectional camera model~\cite{scaramuzza2006toolbox} to keep a fair comparison.}

\begin{figure*}[t]
	\centering
	\subfigure[{Top Trajectory on ID01}]{
		\includegraphics[width=0.3\textwidth]{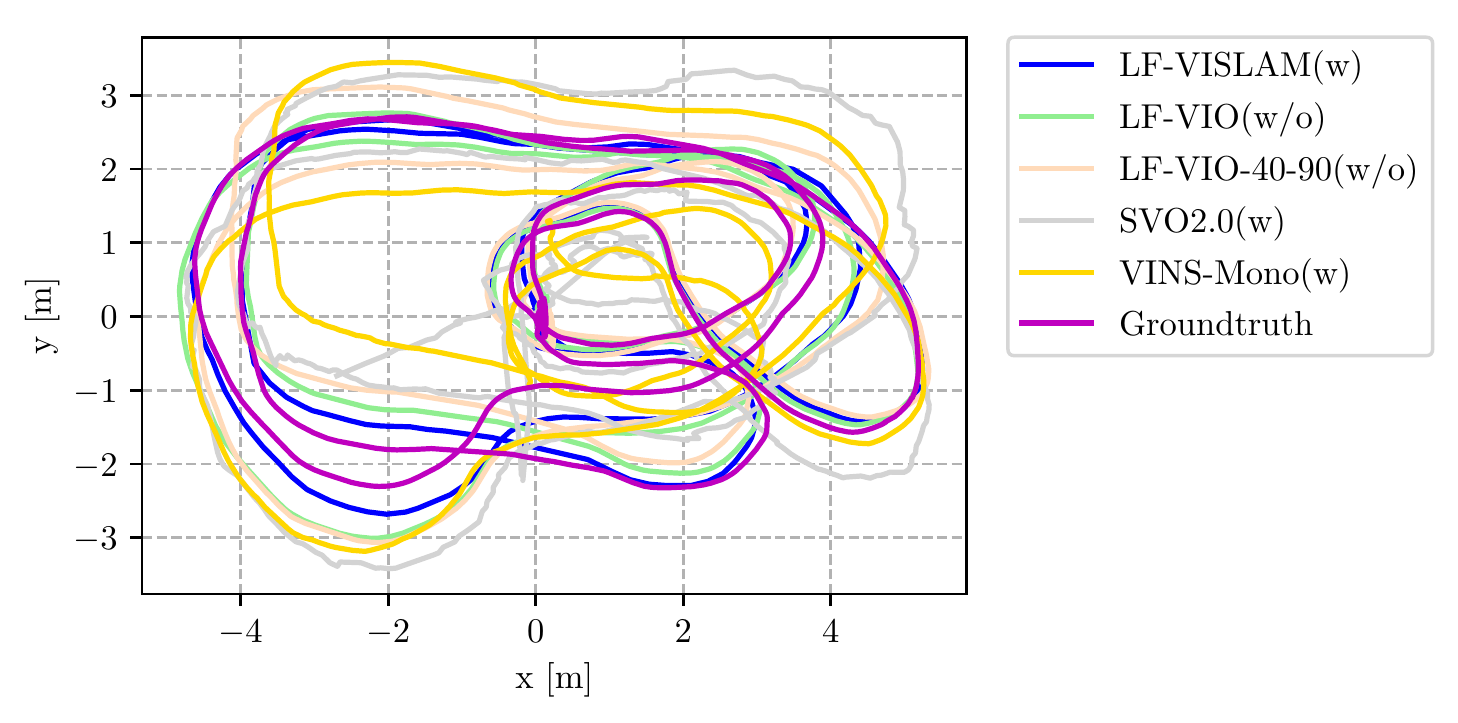}
		\label{fig:mh_01_trajectory_top}
	}
	\subfigure[{Top Trajectory on ID06}]{
		\includegraphics[width=0.3\textwidth]{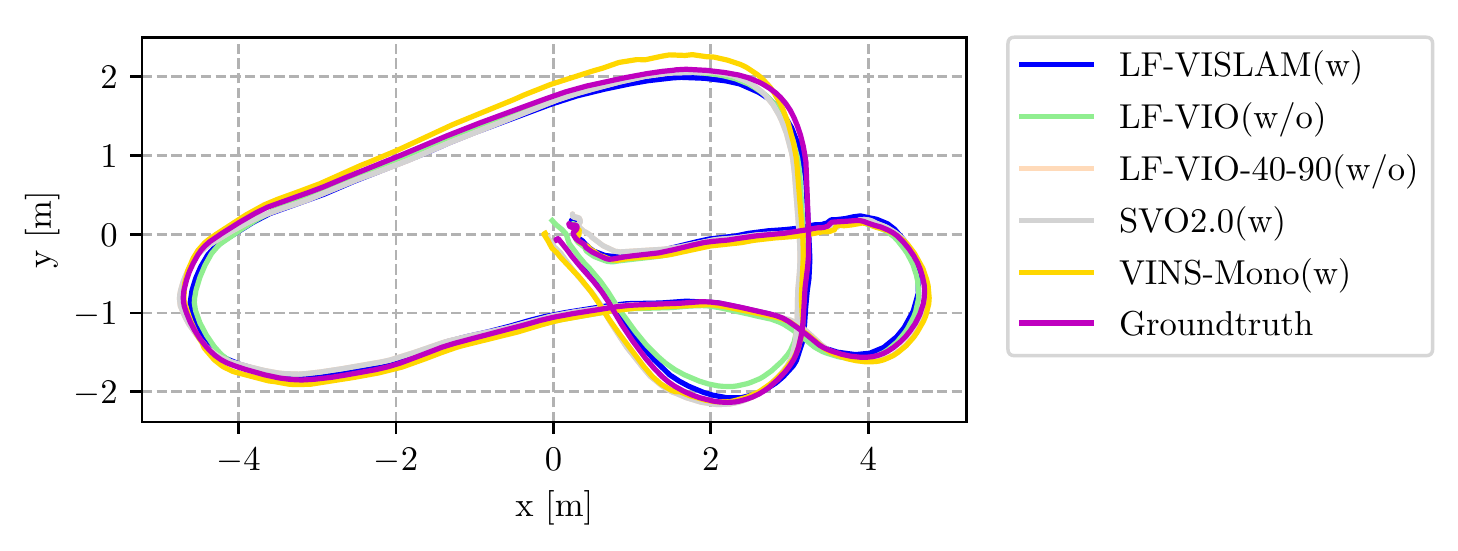}
		\label{fig:mh_06_trajectory_top}
	}
	\subfigure[{Top Trajectory on ID10}]{
		\includegraphics[width=0.3\textwidth]{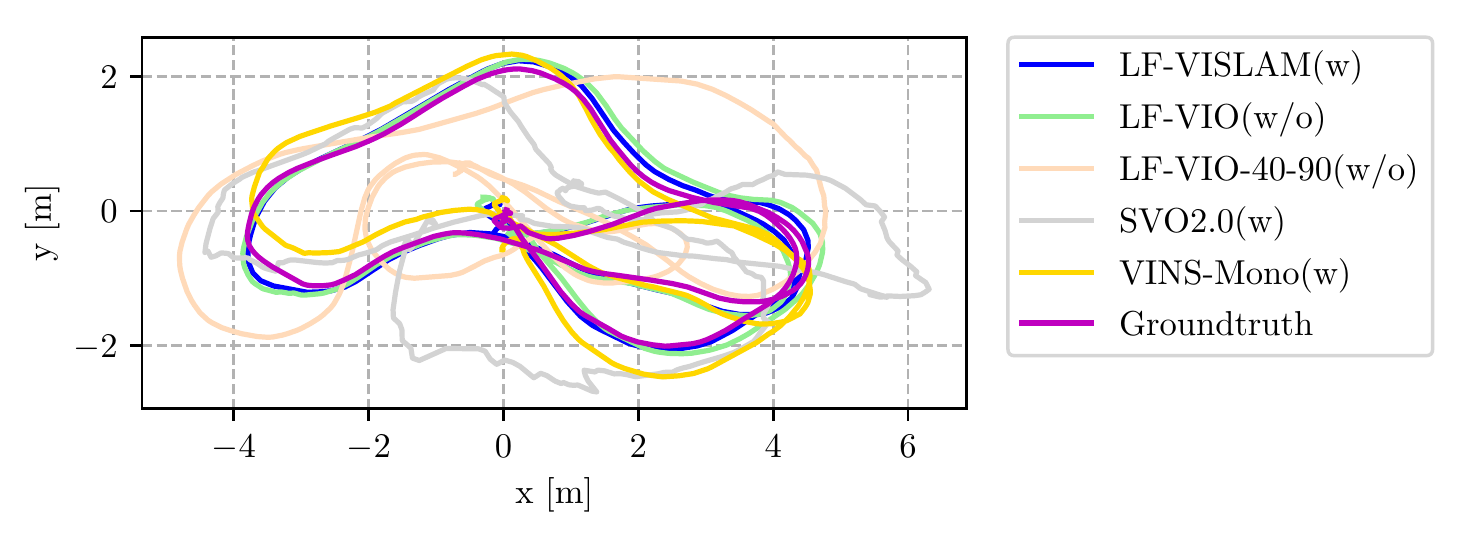}
		\label{fig:mh_11_trajectory_top}
	}
	\subfigure[{Translation and Rotation Error on ID01}]{
		\includegraphics[width=0.3\textwidth]{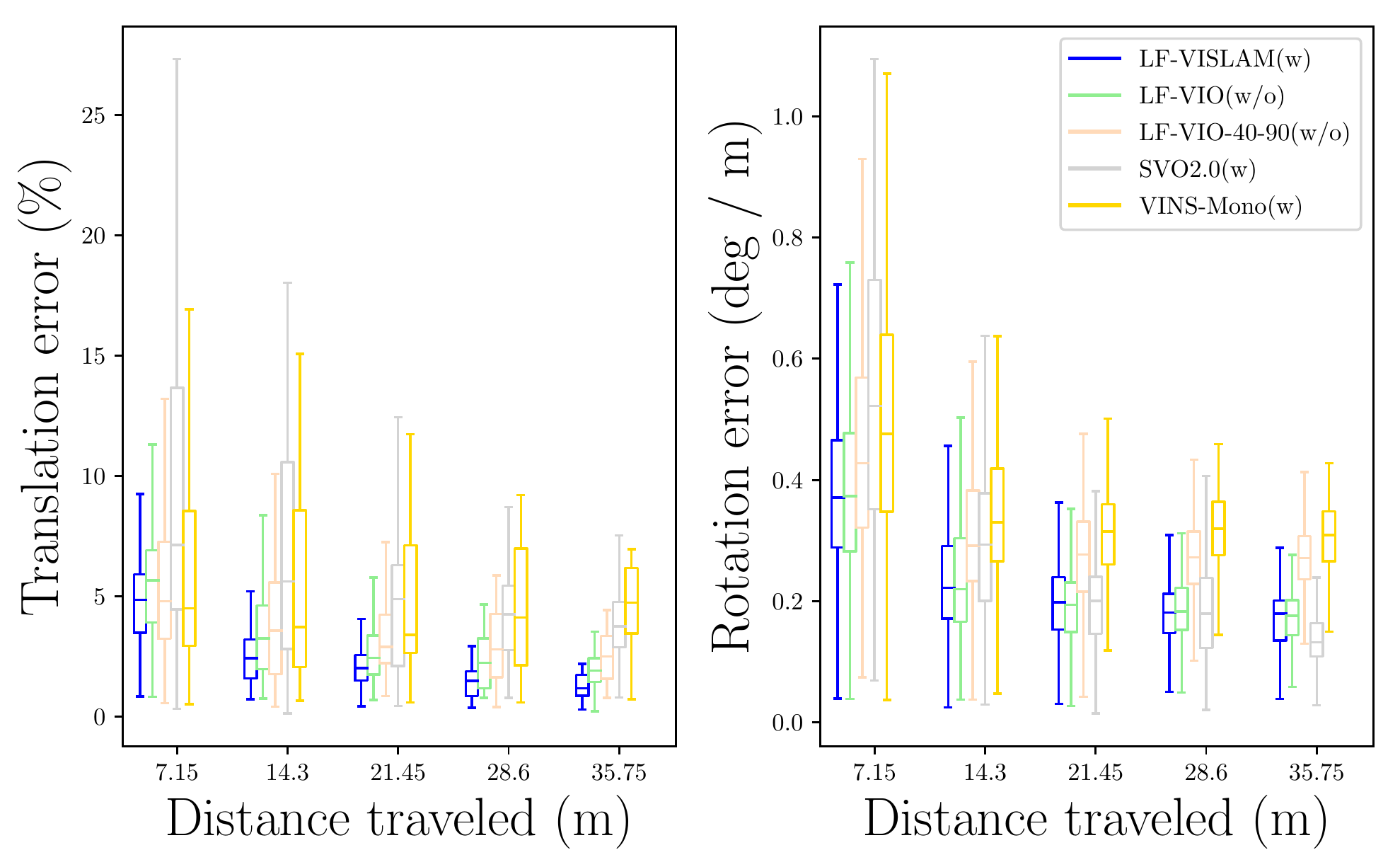}
		\label{fig:mh_01_trans_rot_error}
	}
	\subfigure[{Translation and Rotation Error on ID06}]{
		\includegraphics[width=0.3\textwidth]{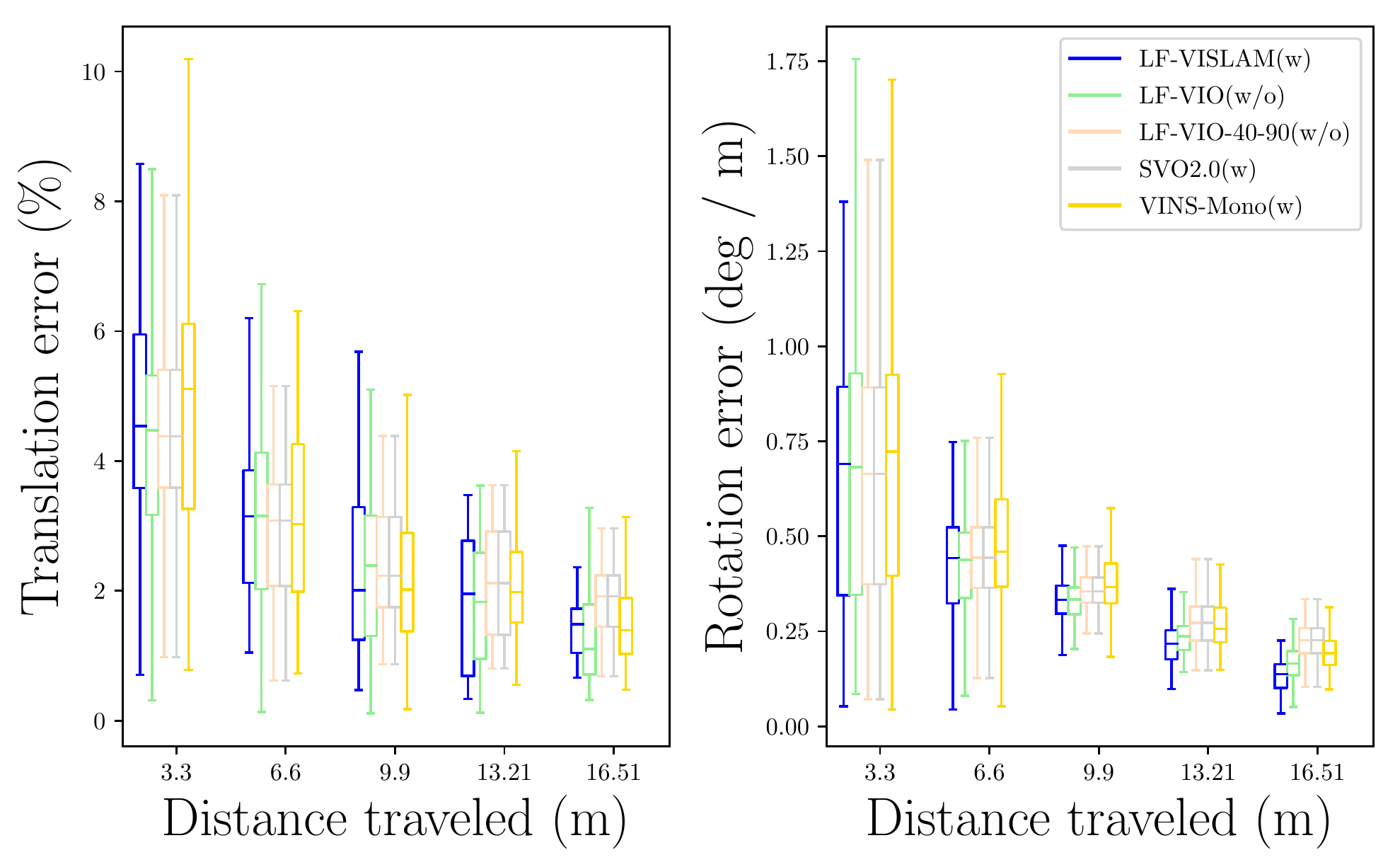}
		\label{fig:mh_06_trans_rot_error}
	}
	\subfigure[{Translation and Rotation Error on ID10}]{
		\includegraphics[width=0.3\textwidth]{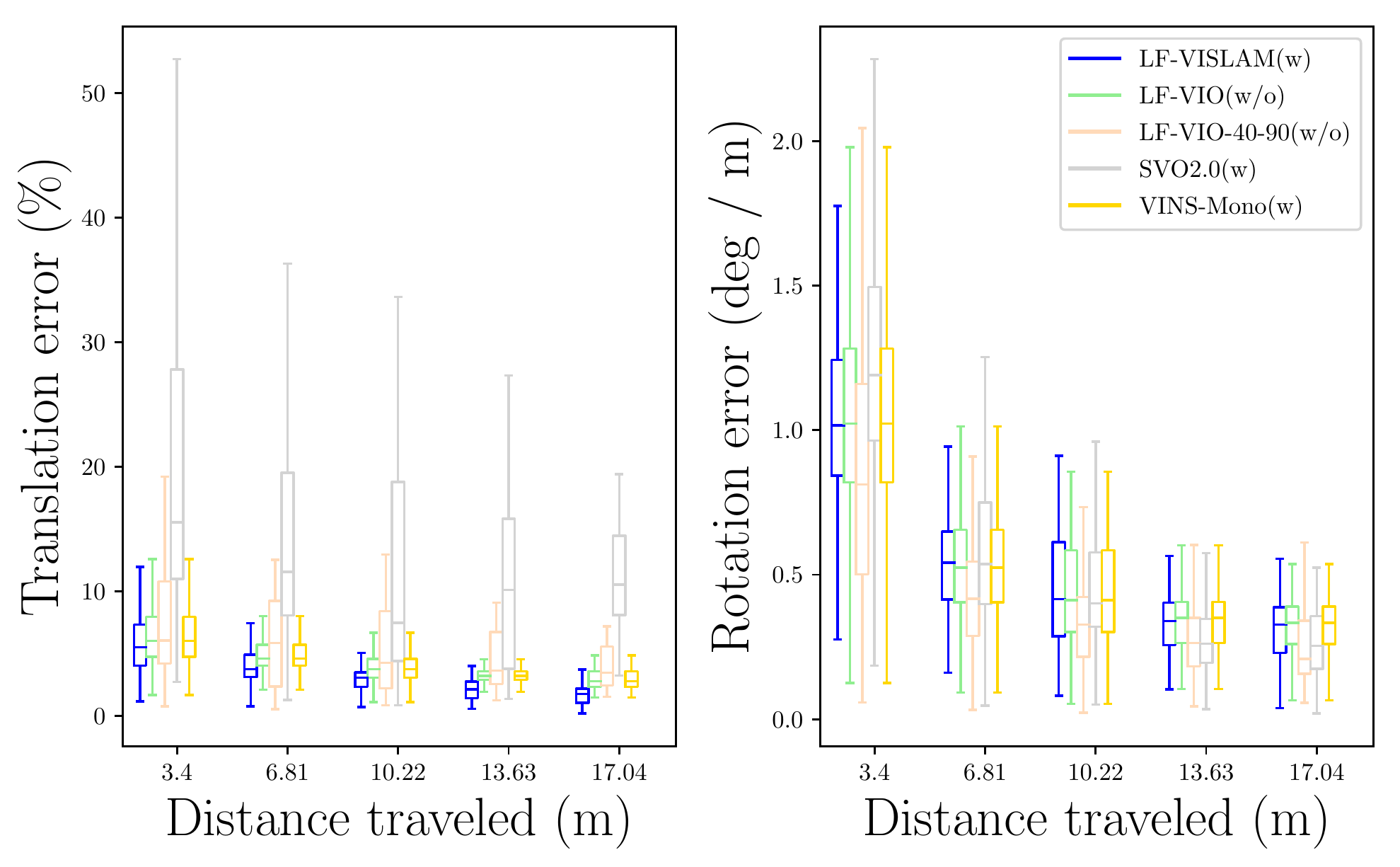}
		\label{fig:mh_11_trans_rot_error}
	}
	\caption{Examples of top trajectories and error analyses on the PALVIO benchmark for different SLAM systems.}
	\label{fig:lfvio-vins-svo}
\end{figure*} 

Due to the characteristics of large-FoV panoramic cameras with a negative imaging plane, VINS and SVO2.0, relying on a representation of $[u,v,1]^T$, can only make use of the positive half-plane image. 
Thus, even if similar images are detected, the useful feature point pairs are still rejected.
For VINS-Mono-Loop, even if its loop closure threads are enabled, it does not trigger the loop-closure optimization with dissimilar images and insufficient feature points.

As shown in Table~\ref{tab:slam}, it can be seen that in all sequences (ID01 to ID10), LF-VISLAM has the highest precision in RPEt and ATE compared to the previous LF-VIO~\cite{wang2022lfvio}, SVO2.0~\cite{forster2014svo}, and VINS-Mono(w)~\cite{qin2018vins}.
Compared with our large-FoV VIO framework LF-VIO, LF-VISLAM significantly elevates the performance.
The effect of our loop closure in ID08 is visualized in Fig.~\ref{fig:loop_image}\textbf{(a)} and IDL02 in Fig.~\ref{fig:loop_image}\textbf{(b)}, which demonstrates that our method can successfully close the loop even at very long distances.
Moreover, LF-VISLAM also has better precision scores in RPEt than LF-VIO and VINS-Mono in all sequences and SVO2.0 in most sequences.
We present Top Trajectory, Translation, and Rotation Errors for sequences ID01, ID06, and ID10, as shown in Fig.~\ref{fig:lfvio-vins-svo}.
Evidently, our method is more accurate and more robust than other algorithms.

\begin{figure}[!t]
    \centering
    \includegraphics[width=0.45\textwidth]{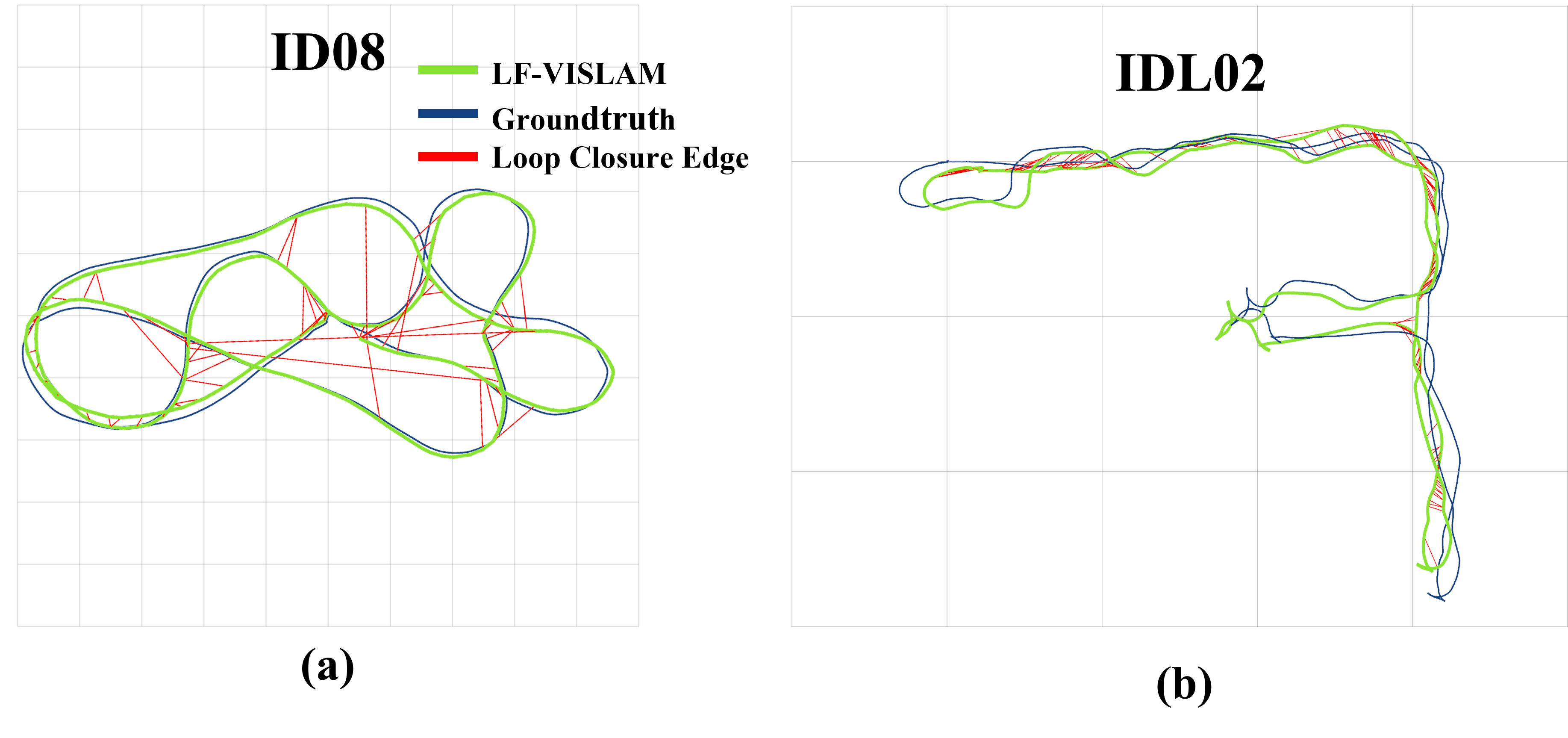}

    \caption{LF-VISLAM's loop closure effect in ID08 on the PALVIO dataset and IDL02. \textbf{(a)} The side length of each gray square is $1m$. \textbf{(b)} The side length of each gray square is $10m$.}
    \label{fig:loop_image}
    
    \vskip-2ex
\end{figure}

\begin{figure}[t!]
	\centering
	\includegraphics[width=1.0\linewidth]{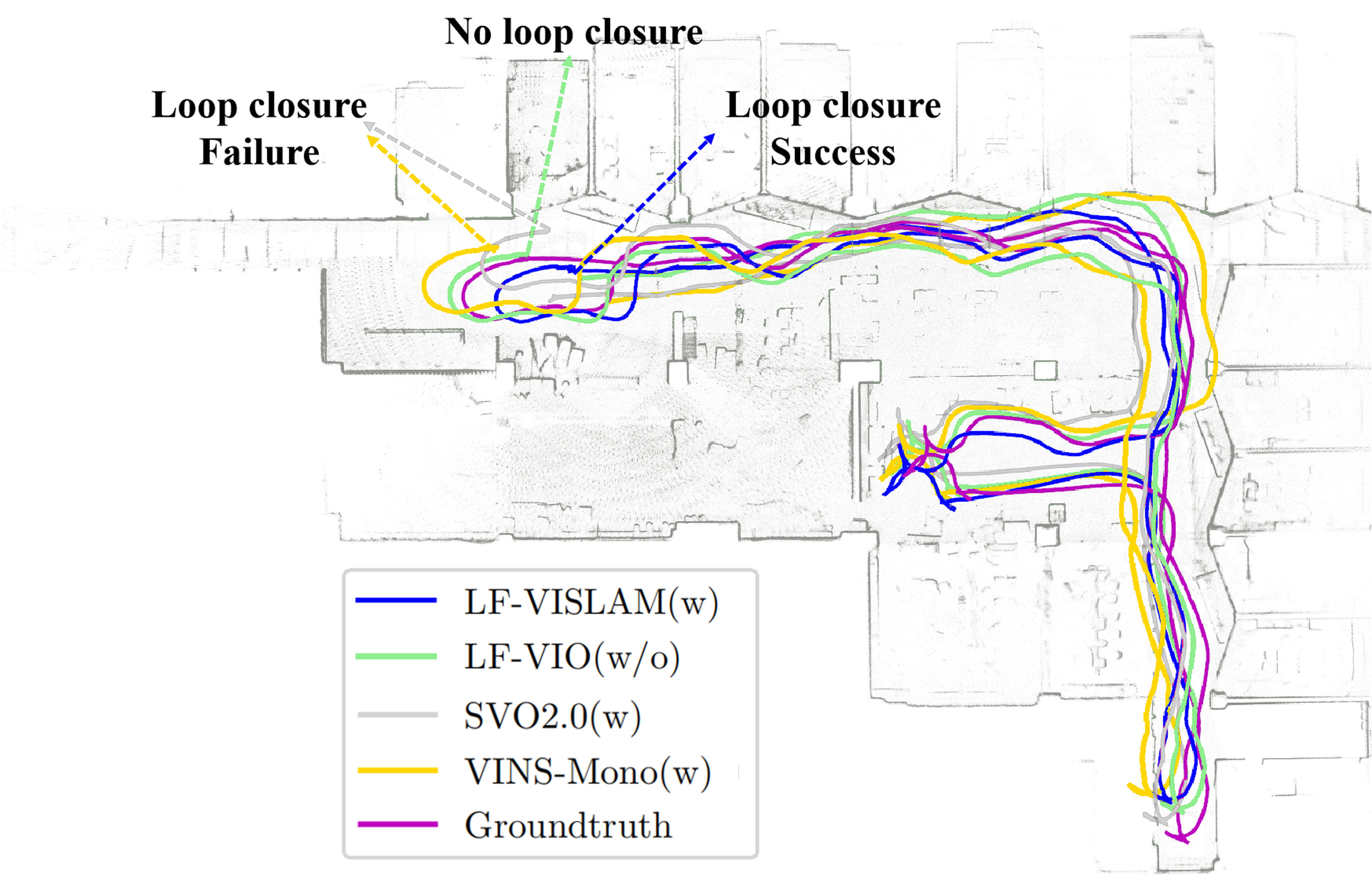}
	\caption{Top trajectory on IDL02 and loop closure effect at the start position.}
	\label{fig:long_traj}
        \vskip-2ex
\end{figure}

\subsection{Comparison on the Long Trajectory Dataset}
\label{exp:long}
To verify the effect of our SLAM algorithm on long trajectories, we record longer trajectory data with a mobile car to evaluate the robustness and accuracy of LF-VISLAM.
Meanwhile, to verify the effectiveness of the loop closure algorithm our path is a closed loop. Due to the long trajectory path, it is not suitable to use motion capture devices.
We use Fast-LIO2~\cite{xu2022fast} with LiDAR SLAM and the closed-loop Scan Context++ algorithm~\cite{kim2021scan} as the ground truth for comparison.
We use the ground vehicle platform (see Fig.~\ref{fig:head_image}\textbf{(b)}) to collect two sequences of data as IDL01 and IDL02, with trajectory lengths of $254.055m$ and $172.431m$, respectively.
Due to the panoramic loop with camera distortions and severe attitude changes, there are a large number of scenes with large differences in yaw angle variations in our dataset.
As shown in Table~\ref{tab:slam_long}, LF-VISLAM clearly outperforms SVO2.0~\cite{forster2014svo} and VINS-Mono~\cite{qin2018vins} in closing the loop under such scenarios with large yaw changes.
In the IDL01 sequence, $327$ loop images are detected, of which $325$ are correct, with a correct rate of $99.39\%$.
In the IDL02 sequence, $103$ loop images are detected, of which $102$ are correct, with a correct rate of $99.03\%$.
In Fig.~\ref{fig:head_image}\textbf{(b)} and Fig.~\ref{fig:long_traj}, we show the top-view trajectory results on IDL01 and IDL02, and it can be clearly seen that our algorithm closes the loop smoothly at the starting point, while SVO2.0 and VINS fail the loop closure. 

\begin{table*}[t]
 \setlength{\tabcolsep}{2.2pt}
	\centering
	\caption{Comparison of SLAM methods on the long trajectory dataset. ``w'': with loop closure, ``w/o'': without loop closure. }
	\label{tab:slam_long}
\renewcommand\arraystretch{1.4}{\setlength{\tabcolsep}{0.15mm}{
\begin{tabular}{lcccccccccccc}
\toprule   
\multicolumn{1}{c|}{\multirow{2}{*}{Sequences}} & \multicolumn{3}{c|}{LF-VISLAM(w)}                                            & \multicolumn{3}{c|}{LF-VIO(w/o)}                                           & \multicolumn{3}{c|}{SVO2.0(w)}                                             & \multicolumn{3}{c}{VINS-Mono(w)}                                   \\
\multicolumn{1}{c|}{}                           & RPEt (\%)            & RPEr (degree/m)      & \multicolumn{1}{c|}{ATE (m)} & RPEt (\%)            & RPEr (degree/m)      & \multicolumn{1}{c|}{ATE (m)} & RPEt (\%)            & RPEr (degree/m)      & \multicolumn{1}{c|}{ATE (m)} & RPEt (\%)            & RPEr (degree/m)      & ATE (m)              \\
\midrule
\midrule
 \multicolumn{1}{c|}{IDL01} &
  \textbf{3.212}        &\textbf{0.116}     &       \multicolumn{1}{c|}{\textbf{2.004}}     &   \underline{3.956}          &0.150      &      \multicolumn{1}{c|}{\underline{2.694}}       &        4.235          &     \underline{0.145}     &\multicolumn{1}{c|}{2.863}  &   4.712    &      0.147                    &3.278      \\
  \multicolumn{1}{c|}{IDL02} &\textbf{3.504}        &\textbf{0.161}     &       \multicolumn{1}{c|}{\textbf{1.190}}     &   \underline{4.225}          &\underline{ 0.168 }      &      \multicolumn{1}{c|}{\underline{1.325}}       &        5.063          &    0.173      &\multicolumn{1}{c|}{1.706}  &   7.725    &      0.172                  &2.849      \\  
\bottomrule          
\end{tabular}}}
\end{table*}

\subsection{Speed Analysis}
\label{exp:speed}

In this subsection, we present the efficiency of our proposed method LF-VISLAM, comprised of the VIO system and the closure threads.
{The elapsed time of LF-VISLAM compared with different SLAM methods are shown in Fig.~\ref{fig:speed_analysis}. The time consumption increase of our algorithm LF-VISLAM is not much compared with VINS. As SVO2.0 uses multi-thread management methods in its processes to reduce the time overhead, its time consumption is the least.}

In our VIO system, if the image-to-map mapping relationship is stored in memory, the image mapping process takes approximately $1{\sim}2ms$. However, this operation is not necessary for the front end of the LF-VISLAM system. Instead, the feature tracking and optical flow computation take $15ms$ and $3ms$ respectively, and the total front-end cost is around $40ms$. The back-end mainly consists of a solver that takes about $30ms$, leading to a total cost of around $60ms$. It is worth noting that the front-end and back-end operate independently. In the loop closure thread, loop detection takes about $3{\sim}5ms$, the DBoW query takes about $10{\sim}20ms$, the improved RANSAC method takes about $10{\sim}15ms$, and loop graph optimization takes about $3{\sim}5ms$. Since the two threads run independently and the running frequency of LF-VISLAM mainly depends on the running speed of the VIO system, it can reach a rate of at least $10Hz$ on a computer with a built-in quad-core Intel i7-8550U processor, which is suitable for real-time applications on mobile navigation agents. We have tested our algorithm LF-VISLAM on the NUC11TNKi5 hardware platform and our algorithm can run in real time. The NUC11TNKi5 is equipped with an Intel Core i5-1135G7 processor running at 2.40GHz with 8 cores.

\subsection{Discussion}
Panoramic imaging has a larger FoV than pinhole imaging, so its perception range is larger, and panoramic visual odometry is more robust to a single under-textured environment such as a white wall. Compared with other algorithms, we make full use of as much information especially on the negative plane as possible to improve the accuracy and robustness of the odometry. Moreover, we use attitude information solved in the VIO system to improve loop closure accuracy. Of course, repeated scenes such as stairs, in the same building may cause loop closure to fail, and additional processing is required such as distance estimation between loop-closure frames to provide constraints for handling such challenging scenarios.

\vskip+2ex

\begin{figure}[t!]
	\centering
	\includegraphics[width=1.0\linewidth]{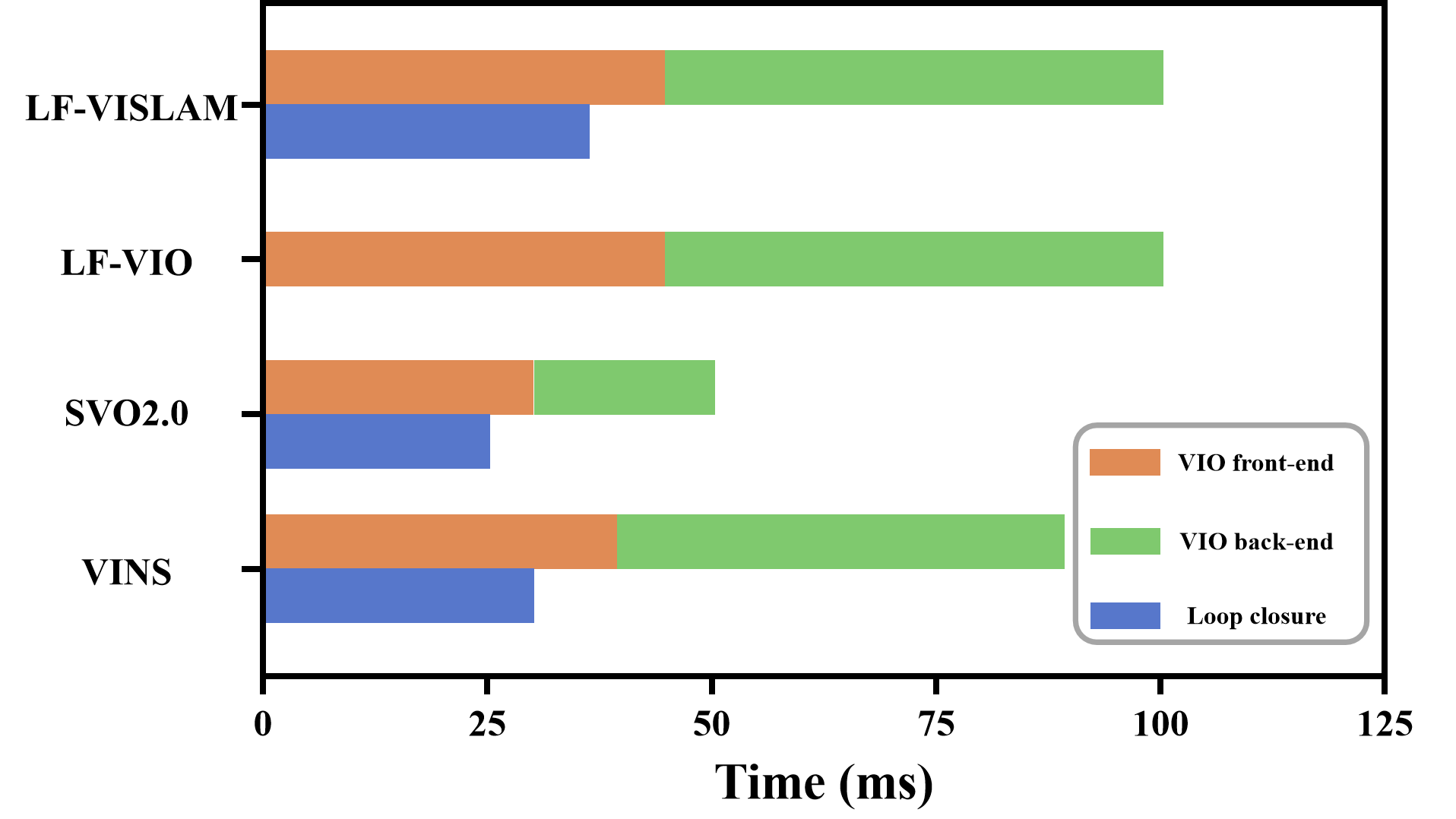}
	\caption{{Elapsed time compared with different SLAM methods.}}
	\label{fig:speed_analysis}
    \vskip-2ex
\end{figure}

\section{Conclusion}

In this article, we have proposed LF-VISLAM, a framework for large-FoV cameras with a negative plane on mobile navigation agents.
LF-VISLAM leverages a unit-length vector for representing feature points to incorporate feature points on the negative plane, which are unused in previous works.
Our large-FoV visual-inertial-odometry further introduces algorithmic adjustments according to the proposed feature representation, and on top of which, an attitude-guided loop closure algorithm is designed to form the entire SLAM system for unleashing the full potential of the ultra-wide $360^\circ$ FoV.

The proposed LF-VISLAM has the following advantages: 
\textbf{1)} It incorporates feature points on the negative plane, leading to a more robust SLAM system;
\textbf{2)} An attitude-guided loop closure algorithm is designed which addresses the challenge posed by large differences in tilt and yaw angles;
\textbf{3)} It has good generalization capacity to long-trajectory sequences and LiDAR-visual-inertial data.

However, there is open room to further improve LF-VISLAM: 
\textbf{1)} The current system is not well compatible with multi-camera systems; 
\textbf{2)} Due to the limited light entering the panoramic annular lens camera, it may not be suitable for low-light environments.
In the future, we intend to incorporate additional sensors, such as multiple large-FoV cameras and wheel speedometers, to enhance the accuracy and robustness of the system. We also aim to explore the use of line features or surface features in the surrounding scene to fully exploit the potential of large-FoV cameras.

\bibliographystyle{IEEEtran}
\bibliography{ref.bib}

\end{document}